\documentclass{article}

\usepackage{arxiv}

\usepackage[utf8]{inputenc} 
\usepackage[T1]{fontenc}    
\usepackage{hyperref}       
\usepackage{url}            
\usepackage{booktabs}       
\usepackage{nicefrac}       
\usepackage{microtype}      
\usepackage{lipsum}

\usepackage{times}

\usepackage[]{algorithm2e}
\usepackage[utf8]{inputenc}
\usepackage[T1]{fontenc}
\usepackage{amsfonts}
\usepackage{amsmath}
\usepackage{amssymb,mathtools}
\usepackage{bbm}

\DeclareMathOperator*{\argmax}{arg\!max}

\newcommand{\ceil}[1]{\left\lceil{#1}\right\rceil}

\usepackage{fullpage,graphicx}

\newcommand {\beq}{\begin{equation}}
\newcommand {\eeq}{\end{equation}}
\newcommand {\bearn}{\begin{eqnarray*}}
\newcommand {\eearn}{\end{eqnarray*}}

\newtheorem{assumption}{Assumption}[section]

\title{Discriminative Learning via Adaptive Questioning}

\author{
  Achal Bassamboo\\
 Northwestern University\\
a-bassamboo@kellogg.northwestern.edu\\
 \And
Vikas Deep\\
Northwestern University\\
vikas.deep@kellogg.northwestern.edu\\
   \And
Sandeep Juneja\\
Tata Institute of Fundamental Research, Mumbai\\
juneja@tifr.res.in\\
\And
Assaf Zeevi\\
Columbia University, New York\\
zeevi@columbia.edu\\
}

\begin{document}
\maketitle

\begin{abstract}
We consider the problem of designing an adaptive sequence of questions that optimally classify a candidate's ability into one of  several categories or discriminative grades. A candidate's ability is modeled as an unknown parameter, which, together with the difficulty of the question asked, determines the likelihood with which s/he is able to answer a question correctly. The learning algorithm is only able to observe these noisy responses to its queries.    We consider this problem from   a fixed confidence-based $\delta$-correct framework, that in our setting seeks to arrive at the correct ability discrimination at the fastest possible rate while guaranteeing that the probability of error is less than a pre-specified and small $\delta$. In this setting  we develop lower bounds on any sequential questioning strategy and develop geometrical insights into the problem structure both from primal and dual formulation. In addition, we 
arrive at algorithms that essentially match these lower bounds. Our key conclusions are that, asymptotically, 
any candidate needs to be asked questions at most at two (candidate ability-specific) levels, although, in a reasonably general framework, questions need to be asked only at a single level. Further, and interestingly, the problem structure facilitates endogenous exploration, so there is no need for a separately designed  exploration stage in the algorithm. 
\end{abstract}


\section{Introduction}
\textbf{Problem overview:}  An evaluator  (the learner or algorithm designer) is tasked with asking an exam taker (the candidate) a sequence of queries
at varying level of difficulty  with  the objective  of classifying some innate property (say, skill level)  of the candidate. 
Specifically, a query may be asked at level 
$x \in {\cal X} \subset \Re^+$ to a candidate with ability $p>0$. The candidate's
response is an indicator function which has a \emph{response function} probability $h(x,p)$ that increases in $p$ and decreases with $x$. The evaluator is unaware of $p$, and the query  $x$ may depend upon 
all the previous questions as well as responses. The grades are specified as 
$J+1$ intervals $([u_i, u_{i+1}), 0 \leq i \leq J)$, each $u_i < u_{i+1}$ and $u_0=0$, $u_{J+1} = \infty$,
and the evaluator's aim is to assign a correct grade to the candidate. 
 
 An overly simplistic and well known instance of such settings are games such as ``20 questions,'' and the problem area as a whole is closely related to various strands of literature that range from sequential hypothesis testing, pure exploration / best arm identification problems, through  psychometric theory / psychological measurement; see further discussion in the (abridged) literature review  below.  

We consider one formulation that will define the learner's objective and  for which we study the complexity of the above problem. This formulation is known as the $\delta$-correct (or PAC) framework. Here the algorithm is unaware of the candidate's ability $p$, and it sequentially and adaptively asks questions to correctly classify the candidate's grade. The aim is to minimize the sample complexity or the expected total number of questions asked, while restricting the probability of wrong grade assignment to a pre-specified $\delta>0$. 
We develop lower bounds that all $\delta$-correct algorithms must satisfy and  propose  $\delta$-correct  algorithms
that match these lower bounds as $\delta \rightarrow 0$  under mild smoothness assumptions on $h$  when  $\mathcal{X}$ is compact. In Section 4, we  develop  deeper structural insights  for a special class of functions $h$  that include the popular logit structure:
\begin{equation} \label{eq:logit001}
h(x,p)= \frac{e^{bp}}{e^{bp}+ e^{ax +c}},
\end{equation}
for non-negative constants $a$ and $b$, and constant $c \in \Re$, as a special case. Logit models are widely used in academic studies as well as in a variety of application settings (see, e.g., \cite{bartroff2008modern}).

\bigskip 
 
 To better appreciate the underlying problem structure,  we need the notion of {\em optimal} question difficulty levels
  for a candidate with ability $p$ in some  interval $(u_i, u_{i+1})$. 
  The $\delta-$correct property of an algorithm implies that it  will select, with probability at least $1-\delta$, 
the hypothesis  that the underlying parameter lies in the interval $[u_i, u_{i+1})$, restricting the probability of  the alternative hypothesis
that the underlying parameter lies in some other grade interval, to within $\delta$.  
This imposes constraints on 
the expected number of times each question may be asked by a $\delta-$correct algorithm  as a function of the unknown $p$.
Such  constraints are typically justified based on change of measure arguments (see, \cite{LaiRobbins85}, \cite{mannor2004sample}) and are distilled further using the transportation inequality developed by \cite{kaufmann2016complexity}. Using this framework, determining efficient lower bounds on the expected number of times each question is asked 
by a $\delta-$correct algorithm, can be modeled as a solution of a max-min optimization problem, again as
a function of $p$.
Now suppose there exists an oracle who knows the value of $p$, and can use it to solve the max-min optimization problem. In particular, the solution to this optimization problem informs the oracle of the {\em optimal} difficulty level of questions to ask, and how often to ask them, as a function of $p$.  These optimal question levels provide insights into the underlying problem structure that an algorithm attempts to learn. Moreover, they can guide a learning algorithm that does not know $p$, and hence does not know the {\em optimal} difficulty level  of questions to ask,  to infer these levels  adaptively  by solving the max-min problem using, say,  a running estimator of $p$ as its proxy.

\textbf{Main contributions.}
 Methodologically, our work  generalizes the analysis in \cite{kaufmann2016complexity}, and the  
transportation inequality they develop  in the finite arm setting, 
  to a continuum of arms. This 
allows us to handle the more realistic setting where question hardness can take values in a continuum. In addition to formulating the 
max-min problem as a function of $p$, 
 we also formulate a dual to 
 the max-min problem. We show that both the primal as well 
the dual formulation are amenable to elegant geometrical interpretations that provide a great deal of structural insights into the optimal solution. 
Our key conclusion is that for any candidate ability level $p$, the optimal solution involves asking questions at  no more than two candidate-ability-specific levels of difficulty. Further, and non-trivially, we show that
under reasonably general conditions, including the setting when $h$ has logit structure,  these questions need to be asked only at a single candidate-ability-specific level of difficulty. These observations that are teased out of the lower bound complexity analysis greatly simplify the design of learning algorithms. 
 In particular, we propose an algorithm whose sample complexity matches the lower bound asymptotically as $\delta \rightarrow 0$.  The proposed algorithm is sequential and at each iteration relies on using the maximum likelihood estimator
of $p$, and plugging that into the lower bound max-min problem to arrive at the next question as well as the stopping time.   To upper bound the sample complexity of the algorithm,   we use a martingale associated with the derivative of the log-likelihood function   and exploit concentration bounds for this martingale combined with  a careful analysis of the maximum likelihood estimation (see, e.g.,  \cite{magureanu2014lipschitz})  to establish the $\delta$-correct property of the proposed 
 algorithm. 

 We develop these algorithms first when ${\cal X}$ is  finite and observe that in this case the max-min lower bound problem is a finite dimensional linear program that can be efficiently solved using {\em lower envelope} based methods. We then extend the algorithm to the setting where ${\cal X}$ is an interval. In the latter case, under the conditions where asking questions at a  single level of difficulty is optimal, we observe that substantially faster  algorithms to solve the lower bound max-min problem may be designed so that the computational time is substantially reduced.  
In the settings where asking questions at  two-levels of difficulty is optimal, if  the problem is additionally constrained so that  only a single level of difficulty question is allowed, (i.e., the oracle is restricted to asking questions of single level of difficulty) we observe that a $\delta-$correct algorithm can still be designed, where again the  lower bound max-min problem is efficiently  solved at each iteration. Further,  we characterize the asymptotic degradation  (as $\delta \rightarrow 0$) in the sample complexity of the algorithm  vis-a-vis the unrestricted settings.


A qualitative structural insight that emerges in certain settings of the problem is a rather intuitive  monotonicity:  under the optimal solution, a candidate with lower ability should be asked easier questions compared to a candidate with higher ability. This monotonicity, in turn,   supports self-correcting exploration in the resulting algorithm (so that a separate exploration phase is not needed). Thus, every time a candidate answers a question correctly, the algorithm's estimate of the candidate's  ability increases, and hence the next question will be harder. Similarly, every time a candidate answers a question incorrectly, the next question will be chosen to be easier.  This effect is more pronounced in the initial stage, thus promoting exploration, while 
it  tapers off as the number of questions asked becomes large.

\textbf{Organization of the paper.} This section concludes with the brief literature review, providing only selected references.  Section \ref{s-lbd}  focuses on a lower bound derivation in the $\delta$-correct framework. Here we present an optimization problem and its dual that characterizes  the lower bound on sample complexity of $\delta-$correct algorithms. Section \ref{s-pac} leverages these ideas to describe a learning algorithm whose theoretical performance is seen to be asymptotically optimal (as $\delta \rightarrow 0$). Section \ref{s-prop} develops elegant  graphical representations of the primal and dual formulations  of the lower bound and  uses specific functional form for the  response function $h$ to draw insights about the various properties that define the complexity of the problem which allows us to further simplify the proposed algorithm. We also show that our proposed algorithm has inbuilt exploration and hence does not require forced exploration. We numerically illustrate this endogenous  exploration characteristic of the proposed algorithm in Section~\ref{s-prop}.  
Proofs are provided in the appendices.

\textbf{Related literature} Our work is related to the area of  psychometric theory which studies the technique of  psychological measurement. Broadly speaking, the paradigm of testing within psychometric theory can be divided into classical theory and modern latent trait theory. The classical theory is based on the assumption that there is a true score and when measured there is an error in measurement. However, this theory lacks a notion of questions characteristics. In other words, classical test theory cannot separate the candidate characteristic and exam/question characteristic: each of these can only be interpreted in the context of the other.
 Modern latent trait theory (also known as item response theory) is based on relationship between individual's performance on a test item/question and test taker's ability. Our work contributes to the latter theory. (For more details on the paradigm see \cite{van2010elements}.)  The literature on modern latent trait includes study of computerized adaptive testing (CAT), where the goal is to classify the test taker into master or non-master category. The questions in CAT is presented to the person is adaptively based on the earlier responses (see \cite{reckase1983procedure}, \cite{lewis1991computerized} and for a review \cite{georgiadou2007review}).

Our work is also linked with hypothesis testing and experimental design. One of the seminal papers in this area is \cite{chernoff1959sequential}. For recent advances on the topic we refer the reader to \cite{naghshvar2013active} and references therein. \cite{bartroff2008modern} uses the approach of hypothesis testing and applies it to the setup of CAT. 

Note that in our setup we are asking questions up to a stopping time, and the final reward is based on the correct classification of the candidate based on the ability. In this sense, our problem is related to the pure exploration problems studied in the multi-armed bandit literature. For instance, finding the best arm, that is, the arm with the highest mean, with above $\delta$-correct type guarantees, is well studied in literature (see, e.g., \cite{mannor2004sample}, \cite{even2006action},
\cite{bubeck2011pure},   \cite{audibert2010best},
 \cite{garivier2016optimal},  \cite{kaufmann2016complexity},    see \cite{juneja2019partition} for generalizations).

\section{Lower bounds for $\delta$-correct algorithms }\label{s-lbd}

Our sequential framework comprises of a candidate and an evaluator, or more aptly, an algorithm. In addition, we are given levels $u_1, u_2, ...u_{J+1}$ that serve as thresholds for grade brackets $[u_i, u_{i+1})$ for $i = 1, \ldots, J$.
An algorithm needs to determine a  candidate's grade, i.e., determine in which bracket $[u_i, u_{i+1})$ the candidate's ability $p \in (\underline{p}, \overline{p}) \subset \Re^+$ lies, with $0< \underline{p} <\overline{p}<\infty$. To do this, the algorithm  adaptively asks a candidate
questions $X_1, X_2,\ldots$ and so on, where each $X_i \in {\cal X} \subset \Re^+$.

We make the following regularity assumption on the response function $h$ and set ${\cal X}$. All the results in the paper are made under this assumption.
\begin{assumption}\label{assum:h_x} 
 $\cal{X}$ is a compact set in $\Re^+$ such that $\min {\cal X}>0$. Further, $h(x,p)\in(0,1)$ is strictly decreasing in its first argument $x$ and strictly increasing in the second argument $p$. In addition, it is continuously differentiable in both  arguments. 
\end{assumption}

Let $I_t$ denote the indicator function of the response to the query at stage $t$: $I_{t}=1$ if question $t$ is correctly answered, else it equals 0.  As indicated earlier, these responses are drawn from a Bernoulli distribution where for difficulty level  $x$ the probability of a correct answer is given by the response function  $h(x,p)$ and the probability of a wrong answer by $1-h(x,p)$.
For $ t \geq 1$, let ${\cal F}_t$ denote the information contained in the $\sigma$-algebra generated by  $((X_k, I_k),  k \leq t)$. 
At stage $t+1$, the algorithm selects a question  $X_{t+1}$ that is measurable with respect to
${\cal F}_t$.  The algorithm stops 
after asking $\tau$ questions, where $\tau$ is a path dependent stopping time with respect to $\{{\cal F}_t \}$. 
At that stage the algorithm  announces the value $i=0,1, \ldots, J$ signifying that the candidate's ability
lies in the $i$th bracket  $[u_{i},u_{i+1})$.



\begin{definition} 
We say that an algorithm is $\delta$-correct  if for any candidate with  ability $p$ in some grade bracket  $(u_i, u_{i+1})$, 
the algorithm stopping time $\tau$ has a finite expectation,  and, for any pre-specified $\delta \in (0,1)$, it
guarantees that the probability of placing the candidate's ability $p$ in  the wrong grade bracket is
bounded from above by $\delta$.
\end{definition} 
As mentioned in the introduction, the  $\delta-$correct property imposes certain restrictions on the algorithm. This can be 
used to arrive at lower bounds on the expected number of required questions. 
This will be done by embedding our candidate-evaluator problem in the evolving pure exploration-based multi armed bandit (MAB) literature.   As mentioned earlier, \cite[Lemma 1]{kaufmann2016complexity} presents a non-asymptotic (in $\delta$) inequality in the MAB context
that gives lower bounds on the expected number of samples generated by each arm  when there are finitely many of such arms.
Proposition~\ref{lemma:Lemma:1} below extends
 this inequality to uncountably many arms and adapts it to  our candidate-evaluator settings.
To this end, let asking a question $x \in {\cal X}$ correspond to
 pulling an arm $x \in {\cal X}$.  Suppose that an  algorithm 
 after $n$ steps generates  data $(X_t, I_t: 1 \leq t \leq n)$.
Further suppose that under probability measure $P$, the candidate ability is
$p \in (u_i, u_{i+1})$ for some $i \leq J$.
Thus,  s/he answers 
question $x$ correctly with probability
$h(x,p)$.  
 For $q,r \in (0,1)$, let
\[
d(q | r) \triangleq  q \log \frac{q}{r} + (1-q) \log \frac{1-q}{1-r}, 
\]
 denote the
Kullback-Leibler divergence between two Bernoulli distributions with means $q$ and $r$, respectively, and  
 $\mathcal{F}_{\tau}$ denotes the $\sigma$ algebra associated with the stopping time $\tau$.
 Suppose that under probability measure, $\tilde{P}$,
 the ability of the candidate equals $u \notin  [u_i, u_{i+1})$.
\begin{proposition}  \label{lemma:Lemma:1}
For an  algorithm with stopping time $\tau$ that is finite in expectation, with probability measures
$P$ and $\tilde{P}$ as above,  
\begin{equation} \label{eqn:01}
E_{P} \left ( \sum_{t=1}^{\tau}  d(h(X_t,p)|h(X_t,u))  \right ) =
\int_{x \in {\cal X}} d(h(x,p)|h(x,u)) dm_{\tau}(x) 
 \geq \sup_{\mathcal{E} \in \mathcal{F}_{\tau}}d(P(\mathcal{E})|
\tilde{P}(\mathcal{E})),
\end{equation} 
where   $m_{\tau}(x) = E_P \left (\sum_{t=1}^{\tau} P(X_t \leq x | {\cal F}_{t-1})\right)$.
\end{proposition}
 
 Equation (\ref{eqn:01}) can be further simplified. For any $\delta-$correct algorithm, for $\delta \in (0,1)$ 
and a stopping time $\tau$ denoting the total number of questions asked,  we have 
\[
\int_{x \in {\cal X}} d(h(x,p)|h(x,u)) dm_{\tau}(x) \geq \log \left ( \frac{1}{2.4 \delta} \right ).
\]
(see \cite{kaufmann2016complexity}).

Let $\overline{x} = \max\{x \in  {\cal X}\}$, and $ \underline{x} = \min\{x \in  {\cal X}\}$. 
Thus, $m_{\tau}(\overline{x})$ denotes $E_P(\tau)$,  the expected number of questions asked
by a $\delta$-correct algorithm corresponding to $\tau$. 
Note that $m_{\tau}(\cdot)$  
is non-negative and non-decreasing  generalized distribution function. It follows that a lower bound on the expected number  of questions 
asked by a $\delta$-correct algorithm is obtained as a $\log \left (\frac{1}{2.4 \delta} \right)$ factor multiplied by the   solution
to the following variational problem (call it ${\bf P1}$): minimize
$m(\bar{x})$  over the space of generalized distribution functions $m(\cdot)$ with support on ${\cal X}$,   such that 
\begin{equation} \label{eqn:eqn1}
\inf_{u \notin [u_i, u_{i+1})} \int_{x \in {\cal X}} d(h(x,p)|h(x,u)) dm(x) 
 \geq 1.
\end{equation}
\subsection{The primal and dual lower bound formulation}
For notational  ease we omit $u_i$, $p$ and $u_{i+1}$ and denote $d(h(x,p)|h(x,u_i))$ by $f_1(x)$ and $d(h(x,p)|h(x,u_{i+1}))$
by $f_2(x)$ (recall that $p \in (u_i, u_{i+1})$). Let $m^*$  denote the optimal solution to ${\bf P1}$. The next result provides primal and dual representation of the asymptotic lower bound of $E_p[\tau]$.

\begin{theorem}[Lower bound primal and dual formulation] \label{prop:prop1}\label{prop:prop2} 
For any $\delta$-correct algorithm with a finite stopping time $\tau$, we have 
$$
\frac{E_P[\tau]}{\log(1/2.4\delta)}\ge m^*,
$$
where $m^*$ equals
\begin{equation} \label{eqn:maxmin}
\left[{\max_{w \in [0,1], x_1, x_2 \in {\cal X}} \,\,\, \min_{j=1, 2}  \left ( w \, f_j(x_1) + (1-w) f_j(x_2) \right )}\right]^{-1}.
\end{equation}
The above equation is referred to as the {\em primal formulation}.
Furthermore, we also have the {\em dual formulation:}
\begin{equation} \label{eqn:maxmin2}
{m^*}=\left[\inf_{{\lambda} \in [0,1]} \,\, \sup_{x \in {\cal X}} \left ( {\lambda} \, f_1(x) 
+ (1-{\lambda}) \, f_2(x)\right)\right]^{-1}.
\end{equation}
\end{theorem}

Recall that both the functions $f_1$ and $f_2$ are parametrized by $p$. To this end, let the optimal solution of \eqref{eqn:maxmin}, be denoted by  $\tilde{w}_1(p)$, $\tilde{x}_1(p)$ and $\tilde{x}_2(p)$.

The proof (in the appendix)  relies on the observation that the infimum in LHS in (\ref{eqn:eqn1}) can only be achieved at $u_i$ or $u_{i+1}$. This allows ${\bf P1}$ to be modeled as a  linear program with two constraints and the number of non-negative variables equal to $|\cal{X}|$. Thus, if $\cal{X}$ is an interval then we have an infinite dimensional linear program. Since, there are only two constraints, we can restrict our search over only two non-negative variables. This implies that questions of only two difficulty levels suffice. The equivalence to (\ref{eqn:maxmin}) follows by a simple normalization of variables so that they sum to 1. The alternative characterization given in \eqref{eqn:maxmin2} of the lower bound is obtained by considering the dual of the linear programming representation of  ${\bf P1 }$. Strong duality follows easily when $|{\cal X}|$ is compact. As with the primal,  (\ref{eqn:maxmin2}) follows by a simple normalization of dual variables so that they sum to 1. 

In the next section, we use the optimization problem in \eqref{eqn:maxmin}  to develop a $\delta$-correct algorithm that estimates $p$ based on the responses to the questions asked by the algorithm and solves the above optimization problem to 
determine the question selection as well as the stopping rule.  Further, in Section~\ref{s-prop} we arrive at  elegant geometric formulations of both the dual and the primal problem to draw insights into the optimal solution, which in turn helps in further simplification of  algorithm design.

\section{An Asymptotically optimal $\delta-$correct algorithm}\label{s-pac}

As discussed in the Introduction, the lower bound analysis in  (\ref{eqn:maxmin})  relies on the knowledge of $p$. This suggests that an algorithm
that at each stage asks questions corresponding to $p$ in  (\ref{eqn:maxmin}) should be optimal.
Such an algorithm 
asks  questions at one or two levels of difficulty depending on $p$, the form of $h(x,p)$, and  ${\cal X}$. The proposed algorithm proceeds sequentially. 
At any stage $t$, an estimator $\hat{p}_t$  of $p$ is used as its  proxy, and questions at the associated optimal level of difficulty are asked. As mentioned before, since the linear program has two constraints, we obtain two levels of hardness for the questions along with the weights as the solution of the min-max problem in Theorem~\ref{prop:prop1}. The proposed algorithm  randomizes between the two levels of hardness of questions appropriately to obtain the lower bound. 
As $\hat{p}_t \rightarrow p$,  the level of difficulty of questions asked are also shown to converge to the optimal levels.
 We further show that the sample complexity of this plug-in algorithm matches $m^*$ as $\delta \rightarrow 0$, thus corroborating the insights obtained from the lower bound analysis. 
 In addition, we prove the $\delta-$ correct property of the proposed algorithm.
 We first develop the algorithm where the set of hardness of question $\cal{X}$ is finite. Later, we note that this
 extends easily to the case where ${\cal X}$ is a compact interval.

Recall that any sequential  algorithm  adaptively asks a candidate,  with  ability $p$, 
questions $X_1, X_2, \ldots$  that are measurable relative to the filtration ${\cal F}_t$ that at stage $t$ is generated by past questions $X_1,\ldots, X_{t-1}$ and  responses  $I_1,\ldots,I_{t-1}$ where $I_t=1$ if question $t$ is correctly answered by the candidate, else it equals 0.  
At any stage $t=1,2,\ldots$, the algorithm also decides whether to stop or not, that is, whether the stopping time $\tau =t$ 
or $\tau >t$. If the algorithm decides to continue, then it must also determine  $X_{t+1}$, the level of difficulty of the next question. If the former,
it announces the value $i=0,1, \ldots, J$ signifying that the candidate's ability is announced to 
lie in the interval $[u_{i},u_{i+1})$. We define ${\bf X}_t=(X_1,\ldots,X_t).$

On the basis of the responses till time $t$, we first wish to estimate the ability of the candidate. This is achieved using the maximum likelihood estimator (MLE). Note that the likelihood of observing data $(I_j: 1 \leq j \leq t)$ when the underlying ability is $p$ and the questions are asked at level ${\bf X}_t$ 
is given by
\[
L({\bf X}_t,p) = \prod_{j=1}^t \left ( h(X_j,p)\right )^{I_j} \left ( 1-h(X_j,p)\right )^{1-I_j}
\]
and the  log-likelihood equals
$
\sum_{j=1}^t I_j \log \left(h(X_j, p)\right) + (1-I_j) \log \left(1-h(X_j, p)\right).
$

We make the following regularity assumption that ensures that we have a unique maximizer for the log-likelihood function. 
\begin{assumption}[The quasi concavity of $\log L({\bf x}_t,p)$] \label{ass2}
We assume that the function $\log L({\bf x}_t,p)$ is strictly quasi concave on $p$ for all vector ${\bf x}_t$ and for all $t \geq 1$.
\end{assumption}
Under Assumption~\ref{ass2}, we are guaranteed a unique maximizer of log-likelihood. We denote this maximizer by $\hat{p}_t$.

\begin{remark} {\em Assumption~\ref{ass2}  should be true for a large class of response functions. We show that 
it holds when,
\begin{equation} \label{eqn:1}
h(x,p) = \frac{g(p)}{g(p)+ k(x)},
\end{equation}
where $g$ and $k$ are strictly increasing differentiable functions. (See Lemma~\ref{uniquenessofmle} and its proof in the appendix.)}
\end{remark}

We first discuss the next question selection rule if the algorithm at any stage decides to continue.  
  Thereafter we discuss the stopping rule. This  does not rely on how the question levels are selected at any stage. 

\bigskip

\noindent \textbf{Question selection rule:} For the algorithm, we find the optimal solution to the min-max problem stated in \eqref{eqn:maxmin} with parameter $\hat{p_t}$ and the feasible set of $x$ is restricted to $\cal{X}$. Let the optimal solution be $\tilde{x}_1(\hat{p_t})$, $\tilde{x}_2(\hat{p_t})$ and $\tilde{w}(\hat{p}_t)$.  
Based on the optimal solution, the next question $X_{t+1}$ is set to $\tilde{x}_1(\hat{p_t})$ with probability $\tilde{w}(\hat{p}_t)$ and  $\tilde{x}_2(\hat{p_t})$ with probability $1-\tilde{w}(\hat{p}_t)$. Of course if $\tilde{x}_1(\hat{p_t})=\tilde{x}_2(\hat{p_t})$ or $\tilde{w}(\hat{p}_t)=0$ or $1$, then there is no randomization needed. 
The question is then asked and the outcome  $I_{t+1}$ observed. 
Then, one checks whether the stopping rule holds or whether the algorithm continues.

\begin{remark} {\em  {Ascertaining  the optimal $\tilde{x}_1(\hat{p}_t)$ $\tilde{x}_2(\hat{p}_t)$ and $\tilde{w}(\hat{p}_t)$ at each $t$ could be computationally expensive. However, one can reduce this computational burden in the following manner.  
Suppose that ${\cal X}= (x_1, x_2, \ldots, x_k)$ where each $0< x_i < x_{i+1}$. Then, the primal lower bound formulation
 leads to a linear program with $k$ variables and two constraints. Let $a_{i}= f_1(x_i)$ and $b_{i}= f_2(x_i)$ for $i=1, \ldots, k$.  The associated primal linear program (See Appendix~\ref{app:algo} for more details) has the form $\min \sum_{i=1}^kt_i$, such that $\sum_{i=1}^ka_{i} t_i \geq 1$,  
 $\sum_{i=1}^kb_{i} t_i \geq 1$, and each $t_i \geq 0$. The corresponding dual linear program has the form
$\max y_1+y_2$, such that $a_{i} y_1 + b_{i} y_2 \leq 1$ for $i = 1, \ldots, k$, and $y_1, y_2 \geq 0$.}

{For positive but arbitrary $\{a_{i}, b_i\}$, this dual is easily solved in $O(k \log k) +O(k)$ time, where $O(k \log k)$ 
term corresponds to sorting the vectors $\{a_{i}, b_i\}$ in the descending order in one of the two components. This relies 
on arriving at a lower envelope, restricted to the positive quadrant, of the lines associated with the $k$ constraints when they are tight.
The proof ideas are simple and essentially well known. They are reproduced in Appendix. If, on the other hand, $\{a_i\}$ or $\{b_i\}$ , have some monotonicity structure. For example, if $\{a_i\}$ first increases with $i$ and then decreases, then sorting the vectors $\{a_{i}, b_i\}$ on the first component involves merging two sorted lists, and is an 
$O(k)$ operation, thus the overall computation time reduces to $O(k)$. Same is true if  $\{a_i\}$ corresponds to a concatenation of
fixed number (independent of $k$) of  monotone sequences.}}
\end{remark}

\bigskip

\noindent \textbf{Stopping rule: } The stopping rule 
corresponds to  the generalized likelihood ratio test adapted to our framework. As is well known, this test considers
the ratio of the likelihood  of observing the data under $\hat{p}_t$ with the likelihood of observing the data under the most likely alternative  hypothesis. The algorithm stops when this ratio is sufficiently large. Suppose that the $\hat{p}_t \in (u_i, u_{i+1})$, so its likelihood equals $L( {\bf X}_t,\hat{p}_t)$.
The likelihood of the most likely alternative hypothesis under Assumption~\ref{ass2} corresponds to $\max (L( {\bf X}_t,u_i), L( {\bf X}_t,u_{i+1})).
$

 The stopping rule corresponds to the log-likelihood ratio, that is,  
\begin{equation}  \label{eqn:911}
\min_{u\in\{u_i,u_{i+1}\}}\left[\sum_{j=1}^t I_j \log \left ( \frac{h(X_j,\hat{p}_t)}{h(X_j,u)} \right ) +
 (1-I_j) \log \left ( \frac{1-h(X_j,\hat{p}_t)}{1-h(X_j,u)} \right )\right].
\end{equation}
exceeding a threshold $\beta(t, \delta)= \log (\frac{c t^{\alpha}}{\delta}(\log(t+1)\log(\frac{1}{\delta}))^{3m+1})$ for the first time, where $m$ is size of the set $\mathcal{X}$, $\alpha=2$, and  $c$ is a  appropriate constant.  This form of function $\beta$ ensures the $\delta-$correctness of the algorithm. 

\begin{algorithm}[h!] 
\SetAlgoLined


\hrulefill \\

\vspace{-1mm}
\textbf{Algorithm 1: $\delta-$correct adaptive questioning for $|{\cal X}|<\infty$} 

\hrulefill

Set $t\leftarrow 1$ and $X_t\leftarrow \underbar{u}$

Ask the question at level $X_t$ and obtain the response $I_t$ (which is 1 if answered correctly 0 otherwise)

Compute $\hat{p}_t$ by maximizing the log likelihood

Set $i$ such that $\hat{p}_t\in[u_{i},u_{i+1})$

 \While{$\min_{u\in\{u_i,u_{i+1}\}}\left[\sum_{j=1}^t I_j \log \left ( \frac{h(X_j,\hat{p}_t)}{h(X_j,u)} \right ) +
 (1-I_j) \log \left ( \frac{1-h(X_j,\hat{p}_t)}{1-h(X_j,u)} \right )\right] \le \beta(t, \delta)$}{

Solve \eqref{eqn:maxmin}  with $\hat{p}_t$ and obtain $\tilde{x}_1(\hat{p}_t)$, $\tilde{x}_2(\hat{p}_t)$ and $\tilde{w}(\hat{p}_t)$ 

Set $X_{t+1}$ to be $\tilde{x}_1(\hat{p}_t)$ w.p. $\tilde{w}(\hat{p}_t)$ and  $\tilde{x}_2(\hat{p}_t)$ otherwise


Increment t by 1

Ask the question at level $X_t$ and obtain the response $I_t$ (which is 1 if answered correctly 0 otherwise)

Compute $\hat{p}_t$ by maximizing the log likelihood

Set $i$ such that $\hat{p}_t\in[u_{i},u_{i+1})$
}
Declare the ability of the candidate lies in the grade bracket $[u_i,u_{i+1}).$ 

\hrulefill
\end{algorithm}

Theorem~\ref{prop:a_o_delta_pac} states our key result. We need the following smoothness conditions:
\begin{assumption}\label{assm:for_ao}
The response function $h(x,p)$ is twice continuously differentiable in $p$. 
Further, $h(x,p)$ satisfies the following conditions for all $x\ge 0$ and $p\ge 0$:
\begin{equation}\label{eqn:supermodular}
\frac{\partial \log(h(x,p))}{\partial x\partial p} \ge  0, \,\,  \frac{\partial \log(1-h(x,p))}{\partial x\partial p} \ge 0.
\end{equation} 
In addition, there exist $k_1,k_2>0$ such that  for $x\in\mathcal{X}$ and $p\in(\underline{p},\overline{p})$, we have
\begin{equation}
\left|\frac{\partial \log(h(x,p))}{\partial p^2}\right| \ge  k_1, \,\,  \left|\frac{\partial \log(1-h(x,p))}{\partial p^2}\right| \ge  k_2.
\end{equation} 
\end{assumption}

\begin{remark}{\em 
Recall that the MLE $\hat{p}_t$ equals $\argmax_{p} L(\mathbf{X}_t,p)$. If, for $I_1$, $I_2$,... $I_t$ fixed,
 $\hat{p}_t$ increases with $\mathbf{X}_t$ then this property can be seen to be equivalent to \eqref{eqn:supermodular}.
The former property may be expected from reasonable response functions $h$ as it essentially requires 
  that the larger $\mathbf{X}_t$ (that is, more difficult questions)  and  responses $I_1$, $I_2$,... $I_t$,
  correspond to a candidate with larger $\hat{p}_t$,  compared to a candidate  who is asked  smaller 
  $\mathbf{X}_t$  (easier questions) and  observed the same responses $I_1$, $I_2$,... $I_t$. 
The logit structured response function given in \eqref{eq:logit001} and the intuitively appealing $h(x,p) = p/(p+x)$ satisfies conditions
  in  Assumption~\ref{assm:for_ao}. In general under mild smoothness conditions the response function given by \eqref{eqn:1} satisfies the Assumption~\ref{assm:for_ao}.
}

\end{remark}

\begin{theorem}\emph{\bf  (Asymptotic optimality of Algorithm 1: Finite ${\cal X}$)}\label{prop:a_o_delta_pac}
Suppose that Assumption~\ref{assm:for_ao} holds,  $p\in (u_i,u_{i+1})$, and $|{\cal X}|<\infty$.  Let $\tau(\delta)$ denote the stopping time for Algorithm 1 describe above. 
 Then,  
\begin{equation}\label{eq-sample_complexity}
\lim_{\delta\rightarrow 0}\frac{{E}_P[\tau(\delta)]}{\log \delta} = -m^* and
\end{equation}
the algorithm is $\delta$-correct, that is, $P\left(\hat{p}_\tau \notin[u_i,u_{i+1})\right)\le \delta.$
\end{theorem}


When ${\cal X}$ is assumed to be an interval, we choose questions from an equi-spaced grid $\cal{X}_\delta$  whose interval size decreases to zero as $\delta \rightarrow 0$ at the rate of order $\log(1/\delta)^{-0.5}$. Further, for the algorithm, we find the optimal solution to the min-max problem stated in (\ref{eqn:maxmin}) with parameter $\hat{p_t}$ and the feasible set of $x$ is restricted to $\cal{X}_\delta$ at $t$. We prove that this modified discretized algorithm is $\delta$-correct and its sample complexity  matches the lower bound developed in (\ref{eqn:maxmin2}) as $\delta \rightarrow 0$. Empirically, we observe that  the performance of the non-discretized algorithm, i.e., solving the max-min problem over the entire ${\cal X}$ is similar to the discretized one.

\begin{corollary} \emph{(\bf Asymptotic optimality of Algorithm 1: ${\cal X}$ is a compact interval)}\label{prop:a_o_delta_pac_a_cont}
Suppose that Assumption~\ref{assm:for_ao} holds, $p\in[u_i,u_{i+1})$, and ${\cal X}$ is a compact interval. Consider Algorithm 1  with ${\cal X}$ set to ${\cal X}_\delta$, and  let $\tau(\delta)$ denote the stopping time. 
 Then,  
\begin{equation}\label{eq-sample_complexity_cont}
\lim_{\delta\rightarrow 0}\frac{{E}_P[\tau(\delta)]}{\log \delta} = -m^*, and
\end{equation}
the algorithm is $\delta$-correct.  That is, $P\left(\hat{p}_\tau \notin[u_i,u_{i+1})\right)\le \delta.$
\end{corollary}

\section{Structural Insights} \label{s-prop}
In this section, we draw insights on the structure of the max-min problems in Theorem~\ref{prop:prop1}, under some restrictions on the response function $h$.
These are useful in designing faster algorithms for solving the max-min problems.  (Recall that Algorithm 1 solves the primal max-min problem at each iteration).  We make the  following regularity assumption.

\begin{assumption}[The quasi concavity] \label{ass1}
${\cal X}= [\underline{x}, \overline{x}]$ for $0<\underline{x} <\overline{x} < \infty$. Furthermore, 
the function  $f(x) \triangleq d(h(x,p)| h(x,u))$, for each $p \ne u$,   is a uni-modal, twice-differentiable, quasi-concave function
of $x \in {\cal X}$. It achieves its maximum  at $x^*(p,u) \in {\cal X}$. For $\underline{x}< x < x^*(p,u)$, $f'(x)>0$ and for $ x^*(p,u)< x< \overline{x}$, $f'(x)<0$. Moreover $ x^*(p,u)$ is an increasing function in $p$ and $u$.
\end{assumption}

This assumption is satisfied by the response function given in \eqref{eqn:1} (See Lemma~\ref{lemma:1} in Appendix~\ref{app:theorem}).
Note that the logit structure for $h(x,p)$ in (\ref{eq:logit001}), is a special case of \eqref{eqn:1}. Further, intuitively appealing, $h(x,p)= p/(p+x)$, is also a special case \eqref{eqn:1}. 
  
\bigskip

Below, we consider  the graphical description of the primal and dual optimization problem under Assumption~\ref{ass1}.
  We are especially interested in identifying conditions under which a single hardness level question is optimal (it solves the associated max-min problem).

\textbf{Graphical description of primal formulation.} As shown in Figure~1(a), under Assumption~\ref{ass1}, the max-min  problem (\ref{eqn:maxmin}) has an elegant graphical description. Fix points $x_1$ and $x_2$. On the graph 
draw a line between the points 
$(x_1, f_1(x_1) $  and $(x_2, f_1(x_2) )$  as well as between $(x_1, f_2(x_1) $  and $(x_2, f_2(x_2) )$.
$\min_{j=i, i+1}  \left ( w \, f_j(x_1)+ (1-w) \, f_j(x_2)) \right )$ as a function of $w$ is the minimum 
of the two lines on the $y$ axis between points $x_1$ and $x_2$. The maximum of this minimum is attained at a point where the two lines intersect. Now look for points $x_1$ and $x_2$ that maximize the height of the corresponding point of intersection.

\begin{figure} \label{fig1}
\begin{center}
\includegraphics[height=0.25\textheight]{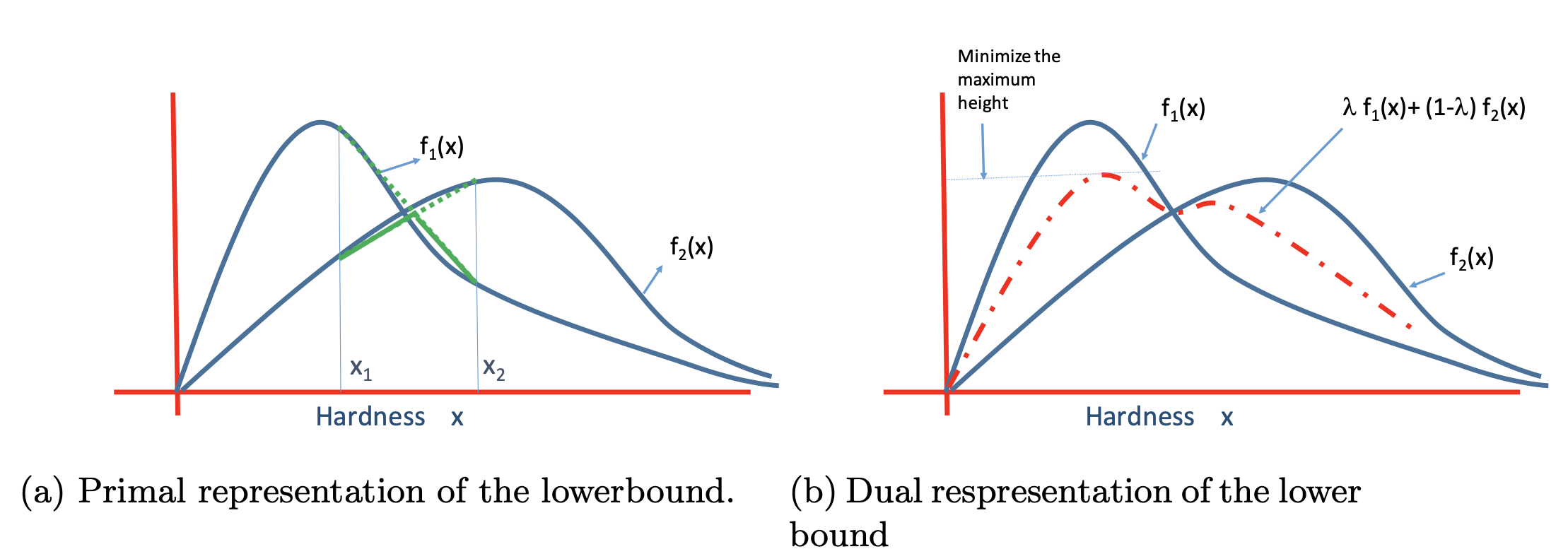}
\end{center}
\caption{Graphical view of the primal and dual representation of the lowerbound.}
\end{figure}

\bigskip

\textbf{Graphical description of dual formulation.} We next move to a geometrical representation of the dual formulation for evaluating $m^*$. For notational ease let us denote $x^*(p,u_i) = x_1^*$ and $x^*(p,u_{i+1}) = x_2^*$ i.e. $ x_1^*$ is the point where $f_1(x)$ achieves its maxima and $ x_2^*$ is the point where $f_2(x)$ achieves its maxima for fixed $p,\,u_i$ and $u_{i+1}$. We know that $x_2^{*}> x_1^{*}$ since $u_{i+1}> u_i$ (See Lemma \ref{lemma:1} in Appendix~\ref{app:theorem}).
As shown in Figure~1(b), (\ref{eqn:maxmin2}) too has a pleasing geometric interpretation: 
The function ${\lambda} \, f_1(x) 
+ (1-{\lambda}) \, f_2(x) $ is a convex combination of the two functions. 
One looks for the convex combination that minimizes
the maximum {\em height} of the resulting function. 
The dual provides new insights into the primal optimal solution. For instance,  if the points $\tilde{x}_1$ and $\tilde{x}_2$ that correspond to the the optimal solution for the primal optimization problem \eqref{eqn:maxmin}, can be seen to correspond to the points where the supremum in \eqref{eqn:maxmin2} can be achieved.  Further,  the dual may be much easier to solve numerically.

\bigskip

 \begin{figure} \label{figvartion}
\begin{center}
\includegraphics[height=0.2\textheight]{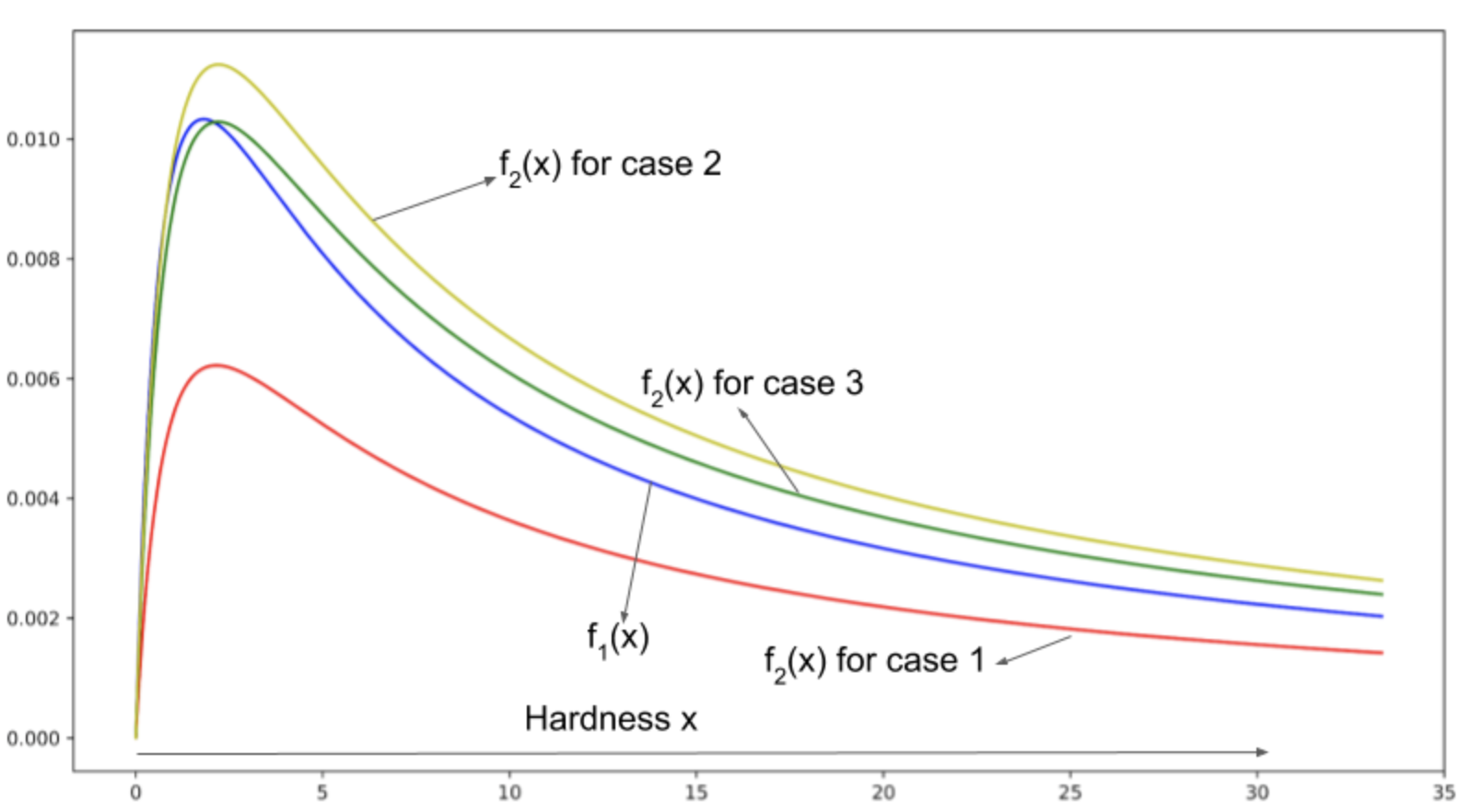}
\end{center}
\caption{A schematic of $f_1(x)$ and  three qualitatively different  cases of   $f_2(x)$.
Here, with fixed $p$ and $u_i$ and varying  $u_{i+1}$, i.e., varying $f_2(x)$. The blue curve  ts $f_1(x)$ while other 3 represent $f_2(x)$ for different cases: {\bf C1)} $ f_1(x^*_2) \geq f_2(x^*_2)$,  {\bf C2)} $f_2(x^*_1) \geq f_1(x^*_1)$, 
and 
 {\bf C3)}  $f_1(x^*_2) < f_2(x^*_2)$ and    $f_2(x^*_1) < f_1(x^*_1)$.}
\end{figure}

\begin{remark}\label{rem:twopoints} {\em 
Observe that due to quasi-concavity of $f_1(x)$ and $f_2(x)$, we may replace
${\cal X}$ with $[x_1^*, x_2^*]$ in the primal as well as the dual formulations in
(\ref{eqn:maxmin}) and (\ref{eqn:maxmin2}), respectively.}
\end{remark}

\subsection{Solving the dual}
 
Under Assumption~\ref{ass1},  based on their maximum values, the pair of functions  $f_1$ and $f_2$
can be segmented into three cases: {\bf C1)} $ f_1(x^*_2) \geq f_2(x^*_2)$,  {\bf C2)} $f_2(x^*_1) \geq f_1(x^*_1)$, 
and 
 {\bf C3)}  $f_1(x^*_2) < f_2(x^*_2)$ and    $f_2(x^*_1) < f_1(x^*_1).$
These three cases are illustrated in Figure~2.
  
\begin{proposition} \label{lemma:lemma101}
Under Assumption~\ref{ass1} and {\bf C1},
\begin{equation} \label{eqn:eqnsj1}
\inf_{\lambda \in [0,1]} \sup_{x \in { [x_1^*, x_2^*]}} \left ( {\lambda} f_1(x) + (1-{\lambda}) f_2(x) 
\right) =  f_2(x^*_2),
\end{equation}
and thus the dual optimal solution corresponds to setting a single question at level $x^*_2$.
This can also be seen to be the optimal solution to the primal. 
Similarly for {\bf C2},
the primal optimal solution corresponds to setting a single question at level $x^*_1$.
\end{proposition}

Proposition~\ref{convex-quasi} below shows conditions under which each
 convex  combination of $f_1(x)$ and $f_2(x)$ is quasi-concave, and thus  aids in arriving at sufficient conditions under which the dual problem (and hence the primal problem) is solved by a single $x$ even for {\bf C3}.

  \begin{proposition}  \label{convex-quasi}
Under Assumption~\ref{ass1},  if the ratio of the derivatives, $\frac{f_1'(x)}{f_2'(x)}$,  is strictly decreasing for $x \in  (x_1^*, x_2^*)$ then $Q(x,\lambda)=$ $\left ( {\lambda} f_1(x) + (1-{\lambda}) f_2(x) 
\right)$ is a quasi concave function in $x$ for $x \in  [x_1^*, x_2^*]$  for all $\lambda \in [0,1]$.
Furthermore,  then under 
{\bf C3},  the $ \bar{x} \in  (x_1^*, x_2^*)$, such that  $f_1(\bar{x})=f_2(\bar{x})$,
 uniquely solves (\ref{eqn:maxmin2}). 
\end{proposition}

In the above proposition, the quasi-concavity of $Q(x,\lambda)$ allows us the use of Sion's minimax theorem to interchange the min-max operations in the dual, so that
\begin{equation} \label{eqn:maxmin_interchange}
\left[\min_{{\lambda} \in [0,1]} \,\, \max_{x \in [x^*_1, x^*_2]} \left ( {\lambda} \, f_1(x) 
+ (1-{\lambda}) \, f_2(x)\right)\right]= \max_{x \in [x^*_1, x^*_2]} \,\, \min_{{\lambda} \in [0,1]} \left ( {\lambda} \, f_1(x) 
+ (1-{\lambda}) \, f_2(x)\right),
\end{equation}
which in turn equals  $\max_{x \in [x^*_1, x^*_2]} \min\{f_1(x),f_2(x)\}= f_1(\bar{x})=f_2(\bar{x})$.  This observation results in an algorithm where  we may solve the dual min-max problem by simply searching for $\bar{x}$  in the interval $[x_1^*,x_2^*]$. 
Recall in Corollary~\ref{prop:a_o_delta_pac_a_cont} and for the associated algorithm we argued that the 
max-min lower bound  problem can be solved in a run time that is linear in  $|\mathcal{X}_\delta|$. 
Further,  we used a grid where $|\mathcal{X}_\delta|= \Theta(\sqrt{-\log \delta})$. 
However, under conditions in Proposition~\ref{convex-quasi}, we are searching for  a single point $\bar{x}$ 
and this can be accomplished in $O(\log |\mathcal{X}_\delta|)= O(\log\log(1/\delta))$ computational time.

\begin{remark} {\em It is also easy to verify that under {\bf C3}, even if the ratio $\frac{f_1'(x)}{f_2'(x)}$ is not strictly decreasing for $x \in  (x_1^*, x_2^*)$,  if the equality in \eqref{eqn:maxmin_interchange} holds then it equals $f_1(\bar{x})$ and single question 
 one hardness level question at $\bar{x}$ is optimal. Thus,  \eqref{eqn:maxmin_interchange} is necessary and sufficient condition for one question to be optimal.  

Under {\bf C3}, if \eqref{eqn:maxmin_interchange} does not hold, so that asking questions at two levels of difficulty
 is optimal, if in our algorithm we use the solution to RHS in \eqref{eqn:maxmin_interchange} with the MLE at any stage 
 plugged into it in place of $p$, then the question to asked is easily ascertained because we are searching for 
 a single point in a closed interval. Further, the resulting  algorithm ( Algorithm 1 when ${\cal X}$ is an interval)
with the caveat that the single point solution is rounded of to the nearest grid-point in ${\cal X}_{\delta}$, 
 will still be $\delta$-correct. This is true as  the $\delta$-correct property stems from the stopping rule which can be applied any set of questions. However, the stopping time $\tau(\delta)$ obtained from such algorithm satisfies
$$
\lim_{\delta\rightarrow 0}\frac{{E}_P[\tau(\delta)]}{\log \delta} = -[
\sup_{x \in {\cal X}} \min\{f_1(x),f_2(x)\}]^{-1}<-m^*=-\left[\inf_{{\lambda} \in [0,1]} \,\, \sup_{x \in {\cal X}} \left ( {\lambda} \, f_1(x) 
+ (1-{\lambda}) \, f_2(x)\right)\right]^{-1}.
$$
Essentially, the algorithm will have an optimality gap that corresponds to the difference in the 
dual when one interchanges the maximum and minimum. Thus, while the algorithm will conduct each iteration efficently, the number of iterations, or equivalently, the sample complexity, would be sub-optimal. }
\end{remark}
 
The following result is somewhat surprising in its generality and is one of the key results of the paper. The proof, while it relies on elementary ideas, is quite complex. 

\begin{theorem}\label{lemma:Lemma:4}
a) \textbf{\em [Single question optimality]}
For   $h(x,p) = \frac{g(p)}{g(p)+k(x)}$ and $p \in (u_i, u_{i+1})$, the ratio  $\frac{f_1^{'}(x)}{f_2^{'}(x)}$ is a strictly decreasing function for  $x \in  (x^*_1, x^*_2)$, and hence  the solution to {\bf P1} is  achieved by a single question level.

b) \textbf{\em [Monotonicity of optimal question in ability]} Further, the unique solution to {\bf P1} denoted by $\tilde{x}(p)$ is monotonically increases with $p$.
\end{theorem}  
\bigskip{}

\noindent\textbf{Numerical illustration of inbuilt exploration.} Recall the monotone property of the solution $\tilde{x}(p)$ to the lower bound problem spelt out in Theorem~\ref{lemma:Lemma:4}. This property ensures that the algorithms outlined above have inbuilt exploration in them. That is, they do not get trapped around any poor estimator of candidate's  ability. To see this heuristically,  suppose
that  a weak candidate with low $p$, due to a streak of good luck, answers unusually large number of  initial questions correctly. This leads to 
an estimator $\hat{p}$  that is higher than $p$. Due to monotonicity of $\tilde{x}(p)$, the questions asked to this candidate become harder. Similarly, a strong candidate, if it answers an unusual number of questions incorrectly and thus has a lower estimator of the candidate's ability, now faces easier questions. 

Below, we test this characteristic of the proposed estimator numerically. 
 We set the our grade levels to the set of integers of the form $3K + 1$ where $K = 0, 1, 2.. .$. Thus the grade levels are $[1,4)$, $[4,7)$, $[7,10)$ and so on. Response function
in the experiments is $h(x, p) = \frac{p}{p+x}$.
We set the ability of the candidate $p=5.5$ so that it  lies in the grade level $[4, 7]$. We run our proposed algorithm, here we can solve the optimization problem explicitly, so we do not discretize the space $\mathcal{X}$.
To check the importance of initial exploration phase, the algorithm is initiated with three different starting question levels: a) \textbf{Easy start}: We begin by asking question an easy question. Specifically, for the first question we set the  hardness $x=2$, b) \textbf{Optimal start}: Here we  compute the optimal hardness of the question given $p$. Specifically, we set the hardness of first question to be around $x=5.96$. This case will not be possible in practice and will serve as a benchmark. c) \textbf{Hard start}: We begin by asking question a question that is harder than the optimal. Specifically, for the first question we set the  hardness $x=10$ 
The resulting Figure~\ref{normal} in Appendix~\ref{s:study} plots an average of $100$ independently generated sample paths for the three setting in terms of the hardness of question over time. We observe that the three sample path comes close to each other almost instantly at the start. Thus, the time taken under the three scenarios to identify the candidate's ability is similar. This  
indicates that the algorithm converges to the optimal hardness level at about the same time, more or less independent of the initial state due to the inbuilt exploration property.

\newpage


\appendix
\section*{Appendix : Proofs of the results in Main Section}\label{app:theorem}
\textbf{Preamble.}
We now present the proofs of the results of the paper in the order they appear in the paper.
\bigskip

\noindent {\bf Proof of Proposition~\ref{lemma:Lemma:1}:}  
Recall that $\mathcal{F}_n$ denotes the $\sigma$-algebra associated with random variables
$(X_i, I_i: 1 \leq i \leq n)$. $m_{\tau}(x) = E_P \left (\sum_{t=1}^{\tau} P(X_t \leq x | {\cal F}_{t-1})\right)$.

Also note that 
\[
Z_n =  \sum_{t=1}^{n} \left (I_t \log \left ( \frac{h(X_t,p)}{h(X_t,u)} \right ) +
(1- I_t)\log \left ( \frac{1-h(X_t,p)}{1-h(X_t,u)} \right ) \right )
\]
 denotes the log-likelihood ratio of $P$ w.r.t. $\tilde{P}$ restricted to ${\cal F}_n$. 
 
We need to show that Equation (18) in \cite{kaufmann2016complexity} holds. The remaining steps are identical to their proof.
This equation corresponds to showing that 
\begin{equation} \label{eqn:appen1}
E_{P} (Z_{\tau}) =
E_P \left ( \sum_{t=1}^{\tau} f(X_i) \right )
=\int_{x \in {\cal X}} f(x) dm_{\tau}(x)
\end{equation}
where recall that $f(x)=d(h(x,p)|h(x,u))$.   Note that $f(\cdot)$ is non-negative, differentiable and bounded.  

The first equality in (\ref{eqn:appen1}) follows simply by conditioning. To see the second equality, observe that
\[
E (f(X_n)|{\cal F}_{n-1}) = \int_{x \in {\cal X}} f'(x) P(X_n\ge x| {\cal F}_{n-1})dx.
\]

Further, we have that 
\[
M_n \triangleq  \sum_{t=1}^n \left ( f(X_t)  - E (f(X_t)|{\cal F}_{t-1}) \right ) 
\]
is a martingale with bounded increments.
By martingale stopping time theorem
\begin{eqnarray}
 E(\sum_{t=1}^{\tau} f(X_t)) = E  \sum_{t=1}^{\tau} E (f(X_t)|{\cal F}_{t-1})
 &=& E \sum_{t=1}^{\tau} \int_{x \in {\cal X}} f'(x) P(X_n\ge x| {\cal F}_{n-1})dx\\
 &=& \int_{x \in {\cal X}}  f'(x) E \sum_{t=1}^{\tau} P(X_n\ge x| {\cal F}_{n-1})dx,
\end{eqnarray}
where the last equality follows from Fubini.  
Further note that using integration by parts and definition of $m_\tau(x)$ we also have 

\[
\int_{x \in {\cal X}} f(x) d m_{\tau}(x) = \int_{x \in {\cal X}} f'(x) E \sum_{t=1}^{\tau} P(X_n\ge x| {\cal F}_{n-1})dx
\]
This completes the proof. 
$\Box$
\vskip 0.4in

\noindent {\bf Proof of Theorem~\ref{prop:prop1} (Primal Representation) }  Since the infimum in LHS in (\ref{eqn:eqn1}) can only be achieved at $u_i$ or $u_{i+1}$,  (\ref{eqn:eqn1}) is equivalent to 
 \begin{equation} \label{eqn:eqn2}
\int_{x \in {\cal X}} f_1(x) dm_{\tau}(x) 
 \geq \log \left (\frac{1}{2.4 \delta} \right),
\end{equation}
and
 \begin{equation} \label{eqn:eqn3}
\int_{x \in {\cal X}} f_2(x) dm_{\tau}(x) 
 \geq \log \left (\frac{1}{2.4 \delta} \right).
\end{equation}

Thus, our optimization problem  is a linear program with uncountably many non-negative variables and, for $m \geq 2$, two constraints.
It follows from semi infinite linear programming theory  (see \cite{lopez2007semi}) that in solving ${\bf P1 }$, it suffices to restrict ourselves to at most two positive variables.
 Thus our optimization problem may be reformulated as:
\begin{eqnarray*}
    \min_{x_1, x_2 \in {\cal X}, m_{x_1}, m_{x_2} \geq 0} &  m_{x_1} +m_{x_2}  &  \\
\mbox{  s.t.  } &  m_{x_1} \, f_1(x_1)  +  m_{x_2} \, f_1(x_2)
  \geq  1 & \label{eqn:eqn7} \\
&   m_{x_1} \, f_2(x_1)  +  m_{x_2} \, f_2(x_2)
  \geq  1. \label{eqn:eqn8} & 
\end{eqnarray*}

By dividing both sides of both the constraints $m_{x_1} +m_{x_2}$, this in turn may be reformulated as
the max-min problem (\ref{eqn:maxmin}).
$\Box$

\bigskip

\vskip 0.1in

\noindent {\bf Proof of Theorem~\ref{prop:prop2}:} (Dual representation)  
  Let 
$\lambda_1, \lambda_2 \geq 0$ such that 
 \begin{equation} \label{eqn:eqn4}
\lambda_1 \, f_1(x) + \lambda_2 \, f_2(x)  \leq 1
\end{equation}
for all $x \in {\cal X}$. The dual problem, call it ${\bf D1}$,  corresponds to maximizing 
$\lambda_1+\lambda_2$ such that (\ref{eqn:eqn4}) holds and $\lambda_1, \lambda_2 \geq 0$.

Observe that for any generalized distribution function $m(\cdot)$ and dual feasible $(\lambda_1, \lambda_2)$,
\[
 \lambda_1 \int_{x \in {\cal X}} f_1(x) dm(x) 
+ \lambda_2 \int_{x \in {\cal X}} f_2(x) dm(x)
\leq m(\bar{x}).
\]
If $m(\cdot)$ is primal feasible in that  (\ref{eqn:eqn2}) and 
(\ref{eqn:eqn3}) hold,  and $(\lambda_1, \lambda_2)$ are  dual feasible,  then weak duality follows, i.e.,
\[
\lambda_1+\lambda_2 \leq m(\bar{x}).
\]
Since $f_1(x)$ and $f_2(x)$ are continuous functions of
$x$, and ${\cal X}$ is compact, strong duality follows.
Thus, there exist
 $\lambda^*_1$, $\lambda^*_2$
that satisfy (\ref{eqn:eqn4}), 
and $(m^*(x): x \in{\cal X})$ that satisfies (\ref{eqn:eqn2}) and 
(\ref{eqn:eqn3}) with
\begin{equation} \label{eqn:eqn61}
\lambda^*_1+ \lambda^*_2 = m^*(\bar{x}).
\end{equation}

Dividing both sides  of (\ref{eqn:eqn4}),
and denoting $\frac{\lambda_1}{\lambda_1+\lambda_2}$ by ${\lambda}$,
solving ${\bf D1}$ is equivalent to solving \begin{equation} \label{eqn:reducex}
\inf_{{\lambda} \in [0,1]} \,\, \sup_{x \in{\cal X}} \left ( {\lambda} \, f_1(x) 
+ (1-{\lambda}) \, f_2(x)\right).
\end{equation}

\begin{remark} {\em
If we can find $\lambda^*_1$, $\lambda^*_2$
that satisfy (\ref{eqn:eqn4}), 
and $(m^*(x): x \in{\cal X})$ that satisfies (\ref{eqn:eqn2}) and 
(\ref{eqn:eqn3})  and (\ref{eqn:eqn61}) holds, then strong duality follows  without
any restriction on ${\cal X}$. Then, $\lambda^*_1$, $\lambda^*_2$
solve ${\bf D1}$ 
and $(m^*(x): x \in{\cal X})$ solves (\ref{eqn:reducex}).
}
\end{remark}

\vskip 0.1in

\bigskip

\noindent\textbf{Proof of  Theorem \ref{prop:a_o_delta_pac} and Corollary \ref{prop:a_o_delta_pac_a_cont}:} 

\bigskip
We provide the proof of  the algorithm described  (with ${\cal X}_\delta$), i.e., Corollary 7,which  subsumes the proof of Theorem \ref{prop:a_o_delta_pac}. Hence we assume that total number of number questions in $\cal{X}$ is $m$, which is increasing with the rate of $(\log\frac{1}{\delta})^{.5}$, for a fixed $\delta$.

\bigskip

We shall provide the proof for a) and b). 

\bigskip
\noindent\textbf{a) Sample Complexity}
To evaluate $E \tau(\delta)$, we divide the state space into a ``good set" and ``bad set." The good set is essentially sample paths on which MLE estimate $\hat{p}_t$ is close to $p$, and the complement set is the bad set. We show that the probability of the bad set is small. On the good set we can show that level of the questions asked are close to optimal solution that one obtains by solving  \eqref{eqn:maxmin2}. Further one shows that the expected stopping time satisfies \eqref{eq-sample_complexity}.

Let $I_j$ be the indicator that the candidate answers the $j^{{th}}$ question correctly. Let $X_j$ denote the level of question $j$. As mentioned earlier after $t^{{th}}$ question, we have the MLE estimator of $p$ denoted by $\hat{p}_t$ that uniquely solves the below equation. We get uniqueness of the MLE due to Assumption  \ref{ass2}. 
\begin{eqnarray}\label{eqn:estiamte_p}
\sum_{j=1}^{t}\frac{I_{j} \,h^{'}(X_j,\hat{p}_t) }{h(X_j,\hat{p}_t)(1-h(X_j,\hat{p}_t))}
=\sum_{j=1}^{t}\frac{h^{'}(X_j,\hat{p}_t)}{1-h(X_j,\hat{p}_t)}.
\end{eqnarray}

After estimating $\hat{p}_t$, we solve the optimization problem (\ref{eqn:maxmin2}) and we get optimal questions that are denoted by $\tilde{x}_1(\hat{p}_t)$ and  $\tilde{x}_2(\hat{p}_t)$ and their proportions $\tilde{w}(\hat{p}_t)$ and $1-\tilde{w}(\hat{p}_t)$, respectively. We then randomize between $\tilde{x}_1(\hat{p}_t)$ and $\tilde{x}_2(\hat{p}_t)$ with proportion $\tilde{w}(\hat{p}_t)$ and then ask that question to the candidate. Without loss of generality, we assume that $\tilde{x}_1(\hat{p}_t) \le \tilde{x}_2(\hat{p}_t)$. Let $J_j$ be the indicator function that $j$th question is of level $\tilde{x}_1(\hat{p}_{j-1})$. We define our good set ${\mathcal{G}_t}$ as follows:
$$
\mathcal{G}_t=\{|\hat{p}_t-p|<\epsilon_1\}\cap\left\{\left|\frac{1}{t}\sum_{j=1}^{t}(J_j-\tilde{w}(p))\right|< \epsilon_2\right\}.
$$

\bigskip{}

\bigskip{}

Later in Appendix in Lemma \ref{lemma:badset}, we show that $\mathbb{P}(\mathcal{G}_t^c)$ is exponentially small. Hence the expected value of the stopping time is governed by the good set. Arguments bounding the contribution to sample complexity from the bad set are similar to those in  \cite{kaufmann2016complexity} and are omitted. 

\bigskip{}

Further note that on the intersection of all the good sets  $\underset{t=1} {\overset{T} \cap}  {\mathcal{G}}_t$, we have that for $i=1,2$
$$
|\tilde{x}_i(\hat{p}_t)-\tilde{x}_i(p)|<\rho_1(\epsilon_1), \,\, |\tilde{w}(\hat{p}_t)-\tilde{w}(p)|<\rho_2(\epsilon_1)
$$
such that $\epsilon_1\rightarrow 0$  implies  that $\rho_1(\epsilon_1)\rightarrow 0$ and $\rho_2(\epsilon_1)\rightarrow 0$. This result follows by continuity of $\tilde{x}_i(\cdot)$ and $\tilde{w}(\cdot)$ which we get by the maximum theorem for the optimization problem in \eqref{eqn:maxmin}.

Observe that,
$$
\min_{u\in\{u_i,u_{i+1}\}}\left[\sum_{j=1}^t I_j \log \left ( \frac{h(X_j,\hat{p}_t)}{h(X_j,u)} \right ) +
 (1-I_j) \log \left ( \frac{1-h(X_j,\hat{p}_t)}{1-h(X_j,u)} \right )\right]
$$
may be re-expressed as
\begin{equation}  \label{eqn:901}
 \min_{u\in\{u_i,u_{i+1}\}} \sum_{j=1}^t \left ( h(X_j,\hat{p}_t) \log \left ( \frac{h(X_j,\hat{p}_t)}{h(X_j,u)} \right ) +
(1-h(X_j,\hat{p}_t)) \log \left ( \frac{1-h(X_j,\hat{p}_t)}{1-h(X_j,u)} \right ) \right )
\end{equation}
\begin{equation}\label{eqn:903}
+ 
\sum_{j=1}^t \left ( \left ( I_j-h(X_j,\hat{p}_t) \right ) \log \left ( \frac{h(X_j,\hat{p}_t)}{h(X_j,u)} \right ) +
 \left(h(X_j,\hat{p}_t)-I_j\right) \log \left ( \frac{1-h(X_j,\hat{p}_t)}{1-h(X_j,u)} \right ) \right ).
\end{equation}

Equation (\ref{eqn:901}) may be expressed  under the good set as 
$$
\min_{u\in\{u_i,u_{i+1}\}}\sum_{j=1}^t \tilde{w}(\hat{p}_j) \left ( h(\tilde{x}_1(\hat{p_j}),\hat{p}_t) \log \left ( \frac{h(\tilde{x}_1(\hat{p_j}),\hat{p}_t)}{h(\tilde{x}_1(\hat{p_j}),u)} \right ) +
(1-h(\tilde{x}_1(\hat{p_j}),\hat{p}_t)) \log \left ( \frac{1-h(\tilde{x}_1(\hat{p_j}),\hat{p}_t)}{1-h(\tilde{x}_1(\hat{p_j}),u)} \right ) \right ) 
$$

\begin{equation}\label{eqn:905}
 + \sum_{j=1}^t (1-\tilde{w}(\hat{p}_j)) \left ( h(\tilde{x}_2(\hat{p_j}),\hat{p}_t) \log \left ( \frac{h(\tilde{x}_2(\hat{p_j}),\hat{p}_t)}{h(\tilde{x}_2(\hat{p_j}),u)} \right ) +
(1-h(\tilde{x}_2(\hat{p_j}),\hat{p}_t)) \log \left ( \frac{1-h(\tilde{x}_2(\hat{p_j}),\hat{p}_t)}{1-h(\tilde{x}_2(\hat{p_j}),u)} \right ) \right )
\end{equation}

\bigskip

Expression in \eqref{eqn:903} can be argued to be relatively small with high probability since $I_j$ is Bernoulli with mean $h(X_j,p)$ and $|p-\hat{p}_t|\le \epsilon_1$.  

Observe that (\ref{eqn:905})  is simply
\[
\min_{u\in\{u_i,u_{i+1}\}}\sum_{j=1}^t \tilde{w}(\hat{p}_j)  \left ( KL (h(\tilde{x}_1(\hat{p}_j),\hat{p}_t) || h(\tilde{x}_1(\hat{p}_j), u)) 
 \right ) + (1-\tilde{w}(\hat{p}_j))\left ( KL (h(\tilde{x}_2(\hat{p}_j),\hat{p}_t) || h(\tilde{x}_2(\hat{p}_j), u)) 
 \right ).
\]
On  the intersection of good sets i.e.  $\underset{t=1} {\overset{T} \cap}  {\mathcal{G}}_t$,  this  is close to
\[
\min_{u\in\{u_i,u_{i+1}\}} t \left [ \tilde{w}(p)KL ((h( \tilde{x}_1(p),p)) || (h(\tilde{x}_1(p), u))) + (1-\tilde{w}(p))KL ((h( \tilde{x}_2(p),p)) || (h(\tilde{x}_2(p), u)))  \right ] = (m^{*})^{-1} t.
\]
The RHS above has to exceed   $\beta(t, \delta)$ and this in turn determines the correct value of $E\tau(\delta)$ yielding property a) in Theorem~\ref{prop:a_o_delta_pac}.

\noindent\textbf{b) $\delta$-correct property}

For the $\delta$-correct property, it suffices to show the following :
\begin{equation}  \label{eqn:921}
P \left (\bigcup_{t}\left\{\sum_{j=1}^t I_j \log \left ( \frac{h(X_j,\hat{p}_t)}{h(X_j,u_i)} \right ) +
 (1-I_j) \log \left ( \frac{1-h(X_j,\hat{p}_t)}{1-h(X_j,u_i)} \right ) > \beta(t, \delta),  
 \hat{p}_t  < u_i \right \}\right) \leq \delta/2. 
\end{equation}

Due to Assumption \ref{ass2}, we get the  quasi concave property of the log-likelihood function. 
The log-likelihood function achieves its maximum value at $\hat{p}_t $ by its definition and since $\hat{p_t}< u_i< p$,  it follows that $\log L(X_t,p) \leq \log L(X_t,u_i)$.

Therefore,

\begin{equation}  \label{eqn:922}
\sum_{j=1}^t I_j \log \left ( \frac{h(X_j,\hat{p}_t)}{h(X_j,u_i)} \right ) +
 (1-I_j) \log \left ( \frac{1-h(X_j,\hat{p}_t)}{1-h(X_j,u_i)} \right ) \le \sum_{j=1}^t I_j \log \left ( \frac{h(X_j,\hat{p}_t)}{h(X_j,p)} \right ) +
 (1-I_j) \log \left ( \frac{1-h(X_j,\hat{p}_t)}{1-h(X_j,p)} \right ).
\end{equation}

Suppose that amongst the first $t$ questions, question $x_k$ is asked
$N_k(t)$ times, and is successfully answered $S_k(t)$ times.

Let $\hat{p}_t(k)$ denote the MLE for $p$ when only questions at level $k$ are considered. It follows that,
\begin{equation}\label{singlemle}
 \frac{S_k(t)}{N_k(t)} = h(x_k, \hat{p}_t(k)).   
\end{equation}

Since $\hat{p_t}$ is the solution of the optimization of  log-likelihood function over all questions asked till time $t$, whereas $(\hat{p}_t(k): 1 \leq k \leq m)$ are the solutions of optimization the likelihood function over the questions asked at level $k$ till time $t$. Hence,  the former  leads to optimal log-likelihood function
value for a more constrained problem compared to the latter.
It follows that log-likelihood function evaluated at $(\hat{p}_t(k): 1 \leq k \leq m)$
 is higher than when evaluated at a MLE $\hat{p}_t$, i.e.,
$$
\sum_{j=1}^t I_j \log \left ( \frac{h(X_j,\hat{p}_t)}{h(X_j,p)} \right ) +
 (1-I_j) \log \left ( \frac{1-h(X_j,\hat{p}_t)}{1-h(X_j,p)} \right )
 =
 $$
 
 $$
\sum_{k=1}^m \left (S_k(t)   \log \left ( \frac{h(x_k,\hat{p}_t)}{h(x_k,p)} \right )
+
 ((N_k(t)-S_k(t))  \log \left ( \frac{1-h(x_k,\hat{p}_t)}{1-h(x_k,p)} \right ) \right ) \leq 
 $$
 
 \begin{equation}  \label{eqn:924}
\sum_{k=1}^m \left (S_k(t)   \log \left ( \frac{h(x_k,\hat{p}_t(k))}{h(x_k,p)} \right )
+
 ((N_k(t)-S_k(t))  \log \left ( \frac{1-h(x_k,\hat{p}_t(k))}{1-h(x_k,p)} \right ) \right ).
\end{equation}

Further, (\ref{eqn:924}) may be expressed as
 \begin{equation}  \label{eqn:925}
\sum_{k=1}^m N_k(t) KL( h(x_k,\hat{p}_t(k)) || h(x_k,p)). 
 \end{equation}

Recall that our aim was to prove (\ref{eqn:921}). Using (\ref{eqn:922}) and (\ref{eqn:925}), we get \eqref{eqn:926} below which in turn bounds from above  (\ref{eqn:921}), the probability of interest. 
 \begin{equation}  \label{eqn:926}
P\left(\bigcup_{t} \left\{\sum_{k=1}^m N_k(t) KL( h(x_k,\hat{p}_t(k)) || h(x_k,p))> \beta(t, \delta)\right \} \right) . 
 \end{equation}
Observe that LHS of (\ref{singlemle}) has expected value $h(x_k,p)$. Hence (\ref{eqn:926}) can be bounded from above by $\delta/2$ using the concentration inequality in \cite{magureanu2014lipschitz} [Lemma~6], by  defining $\beta(t, \delta)= \log (\frac{c t^{2}}{\delta}(\log(t+1)\log(\frac{1}{\delta}))^{3m+1})$ where parameter $c$ is specified later in $\eqref{constantc}$. Thus, 
\[
P\left(\bigcup_{t} \left\{\sum_{k=1}^m N_k(t) KL( h(x_k,\hat{p}_t(k)) || h(x_k,p))> \beta(t, \delta)\right \} \right)
\]

\[
 \leq \sum_{t=1}^\infty e^{(m+1)} \left(\frac{(\beta(t, \delta))^{2}\log(t) }{m} \right)^{m} e^{-\beta(t, \delta)}. 
\]
Substituting the value of $\beta(t, \delta)$
\[
 \leq \sum_{t=1}^\infty \frac{e^{(m+1)}}{m^{m}}\frac { \left (\log(t){\left(\log(ct^{2})+\log(\frac{1}{\delta}) + (3m+1)(\log(\log(t+1)+ \log(\log(\frac{1}{\delta}))))\right)}^2 \right)^{m}}{ct^{2} (\log(t+1))^{3m+1} (\log(\frac{1}{\delta}))^{3m+1}} \delta.\]
To bound the above expression by $\frac{\delta}{2}$, we select c so that,
\[
 2\sum_{t=1}^\infty \frac{e^{(m+1)}}{m^{m}}\frac { \left (\log(t){\left(\log c + 2\log t+\log(\frac{1}{\delta}) + (3m+1)(\log(\log(t+1)+ \log(\log(\frac{1}{\delta}))))\right)}^2 \right)^{m}}{t^{\alpha} (\log(t+1))^{3m+1} (\log(\frac{1}{\delta}))^{3m+1}} \le c.
\]
Recall that $\ceil{\log(\frac{1}{\delta})^{0.5}} = m$.  Hence,
\[
2\sum_{t=1}^\infty \frac{e^{(m+1)}}{m^{7m+2}}\frac { \left (\log(t){\left(3\log t+m^2 + (3m+1)(\log(\log(t+1)+ 2\log(m))\right)}^2 \right)^{m}}{t^{\alpha} (\log(t+1))^{3m+1} (\log(\frac{1}{\delta}))^{3m+1}} \le c
\]

Each term of the above series in the LHS can be bounded from above by 
\begin{equation}\label{eqn:927}
 2\frac{e^{m+1} (\log(t))^{m}({5m^{2}\log(t+1)})^{2m}}{t^{2} (\log(t+1))^{3m+1} m^{(7m+2)}}.
\end{equation}
It follows that,

$$
 2\frac{e^{m+1} (\log(t))^{m}({5m^{2}\log(t+1)})^{2m}}{t^{2} (\log(t+1))^{3m+1} m^{(7m+2)}} \leq 2\frac{e^{m+1} ({5m^{2}})^{2m}}{t^{2} m^{(7m+2)}} .
$$

Since $m$ is a function of $\delta$ and one can observe that,

\[
 \sup_{\delta \in {(0, 1)}} 2\frac{e^{m+1} ({5m^{2}})^{2m}}{t^{2}  m^{(7m+2)}}  \leq \frac{2}{t^2}.
\]
 Hence we can choose  $c = \frac{\pi^2}{3}$  and this completes the proof of (\ref{eqn:921}). 
 
 A similar argument can be used to show:
\begin{equation}  \label{eqn:921a}
P \left (\bigcup_{t}\left\{\sum_{j=1}^t I_j \log \left ( \frac{h(X_j,\hat{p}_t)}{h(X_j,u_i)} \right ) +
 (1-I_j) \log \left ( \frac{1-h(X_j,\hat{p}_t)}{1-h(X_j,u_i)} \right ) > \beta(t, \delta),  
 \hat{p}_t  > u_{i+1} \right \}\right) \leq \delta/2.  
\end{equation}
Combining \eqref{eqn:921} with \eqref{eqn:921a}, we obtain property b) in Theorem~\ref{prop:a_o_delta_pac}.

\bigskip{}

\noindent
{\bf  Proof of Proposition~\ref{lemma:lemma101}:}
Because of Assumption 4.1, $f_1(x)$ and $f_2(x)$ are quasi concave functions and also $x_1^* < x_2^*$ since $x^*(p,u)$ is an increasing function of $u$.
Due to quasi concavity, both $f_1(x)$ and $f_2(x)$ are increasing for $x< x_1^*$, and decreasing for $x > x_2^*$. Hence dual problem (\ref{eqn:maxmin2}) becomes
\begin{equation}
\inf_{{\lambda} \in [0,1]} \,\, \sup_{x \in [x_1^*, x_2^*]} \left ( {\lambda} \, f_1(x) 
+ (1-{\lambda}) \, f_2(x)\right).
\end{equation}
The proof follows by observing that under {\bf C1}, $f_1(x^*_2) \geq f_2(x^*_2)$
implies that
\[
\sup_{x \in [x_1^*, x_2^*]} \left ( {\lambda} f_1(x) + (1-{\lambda}) f_2(x) 
\right) \geq  f_2(x^*_2).
\] 
for every ${\lambda} \in [0,1]$. The equality in (\ref{eqn:eqnsj1}) occurs for 
${\lambda}=0$.
Hence, by strong duality, a primal optimal solution corresponds to setting a single question at level $x^*_2$.
The remaining conclusion similarly follows.

\bigskip{}

\noindent
{\bf  Proof of Proposition~\ref{convex-quasi}:}
Due to Assumption 4.1, we can solve the dual with $x \in [x_1^*, x_2^*]$ as mentioned in the proof of Proposition \ref{lemma:lemma101}. Since $Q(x,1) = f_1(x)$ and $Q(x,0) = f_2(x)$,  we consider $\lambda \in (0,1)$ .

Observe that,
\begin{equation}{\label{nonzero}}
    \frac{\partial Q(x,\lambda)}{\partial x} = \left ( {\lambda} f_1^{'}(x) + (1-{\lambda}) f_2^{'}(x) 
\right).
\end{equation}

To check the quasi concavity of $Q(x, \lambda)$ in $x$, we show that $\frac{\partial Q(x,\lambda)}{\partial x}$ changes the sign only once for $x \in (x_1^*, x_2^*)$.

First we consider $x \in (x^*_1, x^*_2)$.  Rewriting \eqref{nonzero} as,

$$\frac{\partial Q(x,\lambda)}{\partial x} = {\lambda}f_2^{'}(x)\left (  \frac{f_1^{'}(x)}{f_2^{'}(x)} + \frac{1-{\lambda}}{\lambda}
\right) \,\,\,\, \forall x \in (x_1^*, x_2^*),$$
observe that due to quasi concavity of $f_2(x)$, $f_2^{'}(x)$ is always positive for $x \in (x_1^*, x_2^*)$. 
 Thus, at $ x \in (x_1^*, x_2^*)$ where $\frac{\partial Q(x, \lambda)}{\partial x} = 0$, we have,

\begin{equation}{\label{uniquelambda}}
 \frac{f_1^{'}(x)}{f_2^{'}(x)} = -\left ( \frac{1-{\lambda}}{\lambda}
\right).
\end{equation}

Since $\frac{f_1^{'}(x)}{f_2^{'}(x)}$ is a strictly decreasing function for $x \in (x_1^*, x_2^*)$, and at $x \to x_1^*$, $\frac{f_1^{'}(x)}{f_2^{'}(x)} \to 0$ and as $x$ $\to$ $x_2^*$, $\frac{f_1^{'}(x)}{f_2^{'}(x)}$  $\to$  $-\infty$, we conclude that $\frac{\partial Q(x,\lambda)}{\partial x}$ uniquely equals zero at some $\hat{{x}}(\lambda) \in (x_1^*, x_2^*)$  that satisfies (\ref{uniquelambda}).

From (\ref{nonzero}), it follows that at $x = x_1^*, \textrm{ and}\,\,x = x_2^* $, $\frac{\partial Q(x,\lambda)}{\partial x} \neq 0$ for $\lambda \in (0,1)$. 

 It follows that $\frac{\partial Q(x,\lambda)}{\partial x}$ uniquely equals zero for $x \in [x_1^*, x_2^*]$. Also note that $\frac{\partial Q(x,\lambda)}{\partial x} > 0$  for $x > \hat{{x}}(\lambda)$ and 
$\frac{\partial Q(x,\lambda)}{\partial x} < 0$ for $x < \hat{{x}}(\lambda)$
 since $\frac{f_1^{'}(x)}{f_2^{'}(x)}$ is strictly decreasing.
This completes the proof of first part of the lemma.

\bigskip

Now we prove the second part of the lemma. First we prove the existence of $\bar{x}$. Under {\bf C3}, we have $f_1(x^*_1) -f_2(x^*_1) > 0$,
 $f_1(x^*_2) -f_2(x^*_2) < 0$. Since $f_1(x)$ is strictly increasing for $x< x_2^*$ and $f_2(x)$ is strictly increasing for $x> x_1^*$ hence we have
 $f_1'(x) - f_2'(x) <0$ for  $x \in (x^*_1, x^*_2)$.
 Thus, there exists a  unique $\bar{x} \in (x^*_1, x^*_2)$ 
where $f_1(\bar{x})= f_2(\bar{x})$.

Note that  $\left ( {\lambda} \, f_1(x) 
+ (1-{\lambda}) \, f_2(x)\right)$ function is a quasi concave in $x$ for $x \in [x_1^*, x_2^*]$ from  Proposition \ref{convex-quasi}, and linear in $\lambda$ for $\lambda \in [0,1]$.   This allows us to use Sion’s Minimax Theorem to interchange the inf and sup operations to conclude that the solution to (\ref{eqn:maxmin2}) equals

\begin{equation}{\label{soin}}
    \sup_{x \in {[x_1^*, x_2^*]}} \min(f_1(x), f_2(x)).
\end{equation}

It follows that (\ref{soin}) is solved by $\bar{x}$ which satisfies $f_1(\bar{x})= f_2(\bar{x})$, under {\bf C3}.

\bigskip{}

\textbf{Proof of the Theorem~\ref{lemma:Lemma:4}.}
\bigskip{}

We define the following function which will be used in the proof.

\begin{equation}{\label{B(x)}}
 B(x,u_{i},u_{i+1},p) \triangleq   \frac{(p-u_{i} + p \log(\frac{u_{i}}{p}))x + (p^2 + u_{i}p(\log(\frac{u_{i}}{p})-1))}{(p-u_{i+1} + p \log(\frac{u_{i+1}}{p}))x + (p^2 + u_{i+1}p(\log(\frac{u_{i+1}}{p})-1))}.   
\end{equation}

For ease of writing, we suppress the notation $(u_{i},u_{i+1},p)$ in $B(x,u_{i},u_{i+1},p)$ and denote it by $B(x)$. Lemma \ref{lemma:dec_fun} below is useful in proving in Theorem \ref{lemma:Lemma:4}. Its proof is given later.

\begin{lemma}\label{lemma:dec_fun} $B(x)$ defined in (\ref{B(x)}), is a non-negative and strictly decreasing for $x \in (x_1^*, x_2^*)$. Furthermore $\frac{B(x)}{B^{'}(x)}$ is a concave function for $x \in (x_1^*, x_2^*)$.

\end{lemma}

\noindent\textbf{Proof of  Theorem 13 Part (A):}

\bigskip

 First we prove this result for $h(x,p) = \frac{p}{p+x}$ then we extend it for the general family $h(x,p) = \frac{g(p)}{g(p)+k(x)}$.

Observe that for  $h(x,p) = \frac{p}{p+x}$,
\[
  f_1(x) = \log \frac{u_{i}+x}{p+x}  -  \frac{p}{p+x} \log \frac{u_{i}}{p}\textrm{, and } {f_1^{'}(x)} =\frac{(p-u_{i} + p\log(\frac{u_{i}}{p}))x + (p^2 + u_{i}p(\log(\frac{u_{i}}{p})-1))}  {{(p+x)^2}(u_{i}+x)}. 
\]

One can similarly evaluate ${f_2^{'}(x)}$ by replacing $u_i$ by $u_{i+1}$. Hence dividing $f_1^{'}(x)$ by $f_2^{'}(x)$, we get,

\begin{equation}\label{Hfunction}
H(x,u_{i},u_{i+1},p)   \triangleq \frac {f_1^{'}(x)}{f_2^{'}(x)} =  \frac  {{(p-u_{i} + p\log(\frac{u_{i}}{p}))x + (p^2 + u_{i}p(\log(\frac{u_{i}}{p})-1))}(u_{i+1}+x)} {{(p-u_{i+1} + p\log(\frac{u_{i+1}}{p}))x + (p^2 + u_{i+1}p(\log(\frac{u_{i+1}}{p})-1))}(u_{i}+x)}.  
\end{equation}

 \bigskip
  {\bf Analysis of function $H(x,u_{i},u_{i+1},p)$ :}
\bigskip
  
We know that at $x = x_1^*$, $H(x,u_{i},u_{i+1},p)$ is $0$ and as $x$ approaches $x_2^*$, $H(x,u_{i},u_{i+1},p)$  approaches $-\infty$. We also know that $H(x,u_{i},u_{i+1},p)$ is a continuous and differentiable function w.r.t. $x$. Hence to prove that $H(x,u_{i},u_{i+1},p)$ is a strictly decreasing function for $x$  $\in$ $(x_1^*, x_2^* )$, it suffices to show that there is  no root of following equation, for $x$  $\in$ $(x_1^*, x_2^* )$.

$$\frac{\partial H(x,u_{i},u_{i+1},p)}{\partial x} = 0.$$

Using (\ref{B(x)}), we can re-write (\ref{Hfunction}) as
$$H(x,u_{i},u_{i+1},p) = {B(x)} \left(\frac{u_{i+1}+x}{u_{i}+x}\right).$$

Thus,
$$ \frac{\partial H(x,u_{i},u_{i+1},p)}{\partial x} = \left(\frac{(u_{i}-u_{i+1})B(x)}{(u_{i}+x)^{2}}\right) +\frac{dB(x) }{d x}\frac{(u_{i+1}+x)}{(u_{i}+x)}. $$

Our proof relies on the fact $\frac{\partial H(x,u_{i},u_{i+1},p)}{\partial x}$ remains negative in $x \in (x^*_1, x^*_2)$. We show this through a contradiction. 

Suppose $\exists$ $\hat{x}$ that solves
$\frac{\partial H(x,u_{i},u_{i+1},p)}{\partial x} = 0$. Then, $\hat{x}$ must satisfy

 $$\left(\frac{(u_{i}-u_{i+1})B(x) + (B^{'}(x)(u_{i+1}+x)(u_{i}+x))}{(u_{i}+x)^{2}}\right) = 0 .$$
This is equivalent to solving
 $$B^{'}(x)x^{2}+ B^{'}(x)(u_{i+1}+u_{i})x + (u_{i}u_{i+1} B^{'}(x) -(u_{i+1}-u_{i}) B(x)) = 0. $$
Above equation implies that,
\begin{equation} \label{eqn:6}
(2x B^{'}(x) + {(u_{i+1}+u_{i})B^{'}(x)})^2 ={(u_{i+1}-u_{i})B^{'}(x)((u_{i+1}-u_{i})B^{'}(x) + 4B(x))} \textrm{ holds at $x = \hat{x}$}.
\end{equation}
\bigskip

From Lemma \ref{lemma:dec_fun}, we get $B(x)$ $\leq 0$ and $B^{'}(x)<0$ for  $x$  $\in$ $(x_1^*, x_2^*  )$. It follows that $(u_{i+1}-u_{i})B^{'}(x)((u_{i+1}-u_{i})B^{'}(x) + 4B(x)) >0$ and hence we can take the square root of R.H.S of (\ref{eqn:6}).

\bigskip

Recall that our aim is to show 	$\nexists \hat{x}$  that solves (\ref{eqn:6}). We split the problem in two cases based on factorizing (\ref{eqn:6}). Hence $\hat{x}$ if it exists satisfies either case 1 or case 2 below. In the first case we show that $\hat{x} < 0$ hence no solution exists within  $(x_1^*, x_2^*  )$. In case 2 we show that there is no solution to (\ref{eqn:6}).

\bigskip

 {\bf Case 1 :}
 $$ 2xB^{'}(x)+ {(u_{i+1}+u_{i})B^{'}(x)} =+\sqrt{(u_{i+1}-u_{i})B^{'}(x)((u_{i+1}-u_{i})B^{'}(x) + 4B(x))}.$$
 
Since R.H.S. of the equation above is always positive for  $x$  $\in$ $(x_1^*, x_2^*  )$, hence this will lead to the solution $\hat{x}<0$. This implies that no solution exists within $(x_1^*, x_2^*)$.

\bigskip

{\bf Case 2 :} 
   $$ 2xB^{'}(x)+ {(u_{i+1}+u_{i})B^{'}(x)} =-\sqrt{(u_{i+1}-u_{i})B^{'}(x)((u_{i+1}-u_{i})B^{'}(x) + 4B(x))}.$$
  
  Therefore,
  $$ x =\frac{{(u_{i+1}+u_{i})B^{'}(x)}+\sqrt{(u_{i+1}-u_{i})(B^{'}(x))^{2}((u_{i+1}-u_{i}) + \frac{4B(x)}{B^{'}(x)})}}{-2B^{'}(x)}.$$
  
  Since $B^{'}(x)<0$ for $x \in (x_1^*, x_2^*)$ hence  for $x \in (x_1^*, x_2^*)$, above can be written as, 
  
   $$ x =t(x),$$
   where,
   $$t(x) =\frac{{-(u_{i+1}+u_{i})}+\sqrt{(u_{i+1}-u_{i})(u_{i+1}-u_{i}+ \frac{4B(x)}{B^{'}(x)})}}{2} .$$
 \bigskip

By differentiating the above we get,

 $$ t'(x) = \left(\frac{\sqrt{u_{i+1}-u_{i}}\frac{d[\frac{B(x)}{B{'}(x)}]}{dx} }{\sqrt{(u_{i+1}-u_{i}+ \frac{4B(x)}{B^{'}(x)})}}\right) \, \textrm{, and}$$

  $$ t''(x) = \sqrt{u_{i+1}-u_{i}} \left(\frac{\frac{d^{2}[\frac{B(x)}{B{'}(x)}]}{dx^{2}}}{\sqrt{(u_{i+1}-u_{i}+ \frac{4B(x)}{B^{'}(x)})}}+\frac{- 2( \frac{d[\frac{B(x)}{B{'}(x)}]}{dx})^{2}}{[(u_{i+1}-u_{i}+ \frac{4B(x)}{B^{'}(x)})]^{\frac{3}{2}}}\right).$$

We want to prove that $\hat{x} \neq t(\hat{x})$ for $\hat{x} \in (x^*_1, x^*_2)$. We first show that $t^{''}(x) \leq 0 $ for all $x$  $\in$ $(x_1^*, x_2^*  )$. 

Lemma \ref{lemma:dec_fun} implies that  $B(x) \geq 0,\,\, B^{'}(x) < 0 \textrm{ and }\frac{B(x)}{B^{'}(x)} \leq 0$ for $x$  $\in$ $(x_1^*, x_2^*  )$. From this we conclude that $t^{''}(x) \leq 0 $ for $x$  $\in$ $(x_1^*, x_2^*  )$. Hence  $t^{'}(x)$ decreases with x.

From (\ref{LbyL'}), we know that $\frac{B(x)}{B^{'}(x)} \vert_{x = x_1^*} =0$ and the $\frac{d[\frac{B(x)}{B{'}(x)}]}{dx}  \vert_{x = x_1^*} = 1$. Hence we can compute that $t(x_1^{*})= -u_{i}$ and $t^{'}(x_1^*) = 1$ . 

Since $t^{''}(x) \leq 0$ for  $x$  $\in$ $(x_1^*, x_2^*)$, 
$$t^{'}(x) \leq t^{'}(x_1^*).$$   
Substituting the value of $t^{'}(x_1^*)$, it follows that
\begin{equation}{\label{t_dash(x)}}
 t^{'}(x) \leq 1 \, \forall x \, \in (x_1^*, x_2^*). 
\end{equation}

Since  $x_1^{*} > 0$ and  $t(x_1^{*})= -u_{i}$,
\begin{equation}{\label{t(x)}}
  t(x_1^{*}) < x_1^*.
\end{equation}

From (\ref{t_dash(x)} ) and (\ref{t(x)}), it follows that $t(\hat{x})\neq \hat{x}$ for   $\hat{x}$  $\in$ $ (x_1^*, x_2^*  )$ .

 \bigskip

Hence we conclude that $H(x,u_{i},u_{i+1},p)$ is strictly decreasing for $x \in (x_1^*,  x_2^*)$ .

\bigskip
\textbf{
Now we extend this result for $h(x,p) = \frac{g(p)}{g(p)+k(x)}$.
}

\bigskip
 From above we have $\frac{\partial H(x,u_{i},u_{i+1},p)}{\partial x} < 0$ for $x \in (x_1^*,  x_2^*) $ for any given $0<u_{i}<p<u_{i+1}$ when the response function is $\frac{p}{p+x}$. Suppose that for $h(x,p) = \frac{g(p)}{g(p)+k(x)}$, peaks of $f_1(x)$ and $f_2(x)$ are $\bar{x}_1^{*}$ and $\bar{x}_2^{*}$ respectively.

Now when we are replacing $\frac{p}{p+x}$ by $\frac{g(p)}{g(p)+k(x)}$ then one can see that the function $H(x,u_{i},u_{i+1},p)$ becomes $H(k(x), g(u_i), g(u_{i+1}), g(p))$ as given below.

$$H(k(x), g(u_i), g(u_{i+1}), g(p)) = \frac{V(p,u_{i}, x)}{V(p, u_{i+1}, x)} \frac{(g(u_{i+1})+k(x))}{(g(u_{i})+k(x))},$$
$$ \textrm{where, }V(p,u_{i}, x) = (g(p)-g(u_{i}) + g(p)\log(\frac{g(u_{i})}{g(p)}))k(x) + (g(p)^2 + g(u_{i})g(p)(\log(\frac{g(u_{i})}{g(p)})-1)).$$

Differentiating the above,

 $$ \frac{\partial H(k(x), g(u_i), g(u_{i+1}), g(p))}{\partial x} = \frac{\partial H(k(x), g(u_i), g(u_{i+1}), g(p))}{\partial k(x)} (k^{'}(x)).$$
 
Since $H(k(x), g(u_i), g(u_{i+1}), g(p)), \bar{x}_1^{*} \textrm{ and } \bar{x}_2^{*}$ are obtained by the variable change in the definition of  $H(x,u_{i},u_{i+1},p), x_1^* \textrm{ and }x_2^*$,  respectively. It follows that,  

$$ \frac{\partial H(k(x), g(u_i), g(u_{i+1}), g(p))}{\partial k(x)}  < 0 \textrm{ for } x \in ( \bar{x}_1^{*},  \bar{x}_2^{*}),$$
since $k(x)$  is a strictly increasing function. 

\bigskip{}

\noindent\textbf{Proof of Part (B):}

\bigskip

\begin{remark} \label{KL}{\em
Let $f_1(x,p, u_i) = d(h(x,p)|h(x,u_i))$ and $f_2(x,p, u_{i+1}) = d(h(x,p)|h(x,u_i))$ where $ u_{i}< p< u_{i+1}$.
The proof mainly relies on the fact that if we increase the  ability $p$ within the interval $(u_i, u_{i+1})$ then $ f_1(x,p, u_i) $ will increase for each $x$. This is true since increment in $p$ will lead to increase in the $h(x,p)$, and hence $d(h(x,p)|h(x,u_i))$ will also increase. Similarly $f_2(x,p, u_{i+1})$ will decrease if we increase $p$ for each $x$.
}	
\end{remark}

Now we prove the monotonicity of $\tilde{x}(p)$. w.r.t. $p$. Suppose when we increase the ability from $p_1$ to $p_2$, corresponding optimal question's hardness changes from ${\tilde{x}(p_1)}$  to ${\tilde{x}}(p_2)$. As ability ($p$) changes, grade levels i.e. $u_i$ and $u_{i+1}$ can also change hence first we consider the case when $p_1 \in (u_{i},u_{i+1})$ and $p_2 \in (u_{j},u_{j+1})$ where $j \geq{i+1}$.
 \bigskip 
 
 Recall that $x^*(p,u)$ is the the point where function $d(h(x,p)|h(x,u))$ achieves its unique maximum value.  
 From Lemma \ref{lemma:1}, we know $x^*(p,u)$ is an increasing function in both of its arguments and $u_j \geq u_{i+1}$ and $p_2 > p_1 $. Hence using Lemma \ref{lemma:1} and the quasi concavity of $d(h(x,p)|h(x,u))$ , 
 $$\tilde{x}(p_1) \in [x^*(p_1,u_{i}), x^*(p_1,u_{i+1})] \textrm{ and }\tilde{x}(p_2) \in [x^*(p_2,u_{j}), x^*(p_2,u_{j+1})].$$ Again using Lemma \ref{lemma:1}, we get $x^*(p_2,u_{j}) \geq x^*(p_1,u_{i+1})$
 which implies that $\tilde{x}(p_2) \geq{\tilde{x}(p_1)} $.

 \bigskip

 Now consider the case when  $p_1 \in (u_{i},u_{i+1})$ and $p_2 \in (u_{j},u_{j+1})$ where $j = i$. For notational ease we represent $ x_1^*(p_1)= x^*(p_1,u_{i}),\,  x_1^*(p_2)= x^*(p_2,u_{i}),\,  x_2^*(p_1)= x^*(p_1,u_{i+1}) \textrm{ and } x_2^*(p_2)= x^*(p_2,u_{i+1}).$ 
 
Using Lemma \ref{lemma:1} and quasi concavity of $d(h(x,p)|h(x,u))$, 
 
 \begin{equation}\label{monotone}
 \tilde{x}(p_1) \in [x_1^*(p_1), x_2^*(p_1)] \textrm{ and }\tilde{x}(p_2) \in [x_1^*(p_2), x_2^*(p_2)].    
 \end{equation}

 Since grade levels are fixed,  for ease of writing,  denote $f_1(x,p,u_i)$ and $f_2(x,p,u_{i+1})$ by $f_1(x,p)$ and $f_2(x,p)$. Depending upon the structure of functions $f_1(x,p)$ and $f_2(x,p)$ for $p = p_1$ and $p = p_2$, as defined in the main section, we have 9 cases. We represent these cases by $A_{ij}$ where $i= 1,2,3$ and $j=1,2,3$. For an example $A_{12}$ represents $f_1(x,p_1)$ satisfies the condition in {\bf (C1)} and $f_2(x,p_1) $  satisfies the condition in {\bf (C2)}. We prove each of them case by case.

\bigskip
{\bf Cases $A_{21}$, $A_{22}$, $A_{23}$ :} 

 \bigskip

Under {\bf (C2)} for $p = p_1$, from Proposition~\ref{lemma:lemma101}, we get ${\tilde{x}}(p_1)= x^{*}_1(p_1)$. From Lemma \ref{lemma:1} we get $x_1^*(p_2) \geq x^{*}_1(p_1)$. Hence,

$${\tilde{x}}(p_2) \geq x_1^*(p_2) \geq {\tilde{x}}(p_1)$$
 First inequality follows from (\ref{monotone}).

 \bigskip

{\bf Case $A_{31}$ :}  

 \bigskip

 Under this case, using Proposition~\ref{lemma:lemma101}, we get ${\tilde{x}}(p_2) = x_2^*(p_2)$. Using Lemma \ref{lemma:1}, we get $x_2^*(p_2) \geq x_2^*(p_1) $.
Hence,

 $${\tilde{x}}(p_2) =x_2^*(p_2) \geq x^{*}_2(p_1) \geq {\tilde{x}}(p_1).$$ Last inequality follows from (\ref{monotone}).

\bigskip

 {\bf Cases $A_{32}$, $A_{33}$:} 
 
  \bigskip
 
We divide the proof in two parts. First part is when $x_1^{*}(p_2) \geq \tilde{x}(p_1)$ then we can conclude that ${\tilde{x}}(p_2) \geq \tilde{x}(p_1)$ from (\ref{monotone}).

Consider the other case ,i.e., $x_1^*(p_2)< \tilde{x}(p_1)$.
First we show that (\ref{eqn:113}) holds under this part.  We also use this later in the proof.
\bigskip

\begin{equation}  \label{eqn:113}
f_1(x,p_2)>  f_2(x,p_2) \,\forall \, x \in [x_1^{*}(p_2),\tilde{x}(p_1)).
\end{equation}

From the definition of $ \tilde{x}(p_1)$,

$$ f_1(x,p_1) > f_2(x,p_1)  \,\forall \, x \in [x_1^{*}(p_1),\tilde{x}(p_1)).  $$

From Lemma \ref{lemma:1}, we know $x_1^{*}(p_2) \geq x_1^{*}(p_1)$. It follows that,

$$ f_1(x,p_1) > f_2(x,p_1)  \,\forall \, x\in [x_1^{*}(p_2),\tilde{x}(p_1)).  $$

From Remark \ref{KL}, we know that $ f_1(x,p_2) > f_1(x,p_1) \textrm{ and} f_2(x,p_2)< f_2(x,p_1) \,\forall \, x$. Hence,

$$f_1(x,p_2)>  f_1(x,p_1) > f_2(x,p_1) >  f_2(x,p_2) \textrm{ for all } x \in [x_1^{*}(p_2),\tilde{x}(p_1)).$$

Thus,  (\ref{eqn:113}) follows.

\bigskip{}

Now we show that $x_1^*(p_2)< \tilde{x}(p_1)$  is not feasible.

 Under $A_{32}$,  $f_2(x_1^{*}(p_2), p_2) \geq f_1(x_1^{*}(p_2), p_2)$ which violates (\ref{eqn:113}). Hence,
   $x_1^*(p_2)< \tilde{x}(p_1)$ cannot be true. 

\bigskip{}

Under $A_{33}$, we get $f_1({\tilde{x}}(p_2), p_2)=f_2({\tilde{x}}(p_2), p_2)$ using Theorem~\ref{lemma:Lemma:4} and we get ${\tilde{x}}(p_2) \geq x_1^{*}(p_2) $ from (\ref{monotone}) which violates (\ref{eqn:113}). Hence we can conclude that this case ,i.e., $x_1^*(p_2)< \tilde{x}(p_1)$ can not be true.

\bigskip

{\bf Cases $A_{12}$, $A_{13}$:}

We divide the proof in two cases. First corresponds to  $x_1^{*}(p_2) \geq \tilde{x}(p_1)$.  Here we   conclude that ${\tilde{x}}(p_2) \geq \tilde{x}(p_1)$ from (\ref{monotone}).

Now consider  $x_1^*(p_2)< \tilde{x}(p_1)$.

First we show that (\ref{eqn:114})  holds. We use this later in the proof.

\begin{equation} \label{eqn:114}
f_1(x,p_2) > f_2(x,p_2)  \,\forall \, x \in [x_1^{*}(p_2),{\tilde{x}}(p_1) ]
\end{equation}
First observe that $f_1(x,p_1) > f_2(x,p_1)$ $ \,\forall \, x$ $\in [x_1^{*}(p_1),x_2^{*}(p_1) ]$ since  $f_1(x,p_1)$ and $f_2(x,p_1)$ satisfy the conditions in \textbf{(C1)}.

From Remark \ref{KL} we get,
$$ f_1(x,p_2) > f_1(x,p_1) \textrm{ and } f_2(x,p_2)< f_2(x,p_1) \,\, \forall x.$$

Hence,
$$f_1(x,p_2) > f_2(x,p_2)  \,\forall \, x \in [x_1^{*}(p_1),x_2^{*}(p_1) ].$$

From Lemma \ref{lemma:1}, we know $x_1^{*}(p_2) \geq x_1^{*}(p_1)$. It follows that,

$$f_1(x,p_2) > f_2(x,p_2)  \,\forall \, x \in [x_1^{*}(p_2),x_2^{*}(p_1) ].$$
Since under \textbf{(C1)} for $p= p_1$, using Proposition \ref{lemma:lemma101}, we get ${\tilde{x}}(p_1)= x^{*}_2(p_1)$. Thus,  (\ref{eqn:114}) follows.

\bigskip

Under $A_{13}$, we get $f_1({\tilde{x}}(p_2), p_2)=f_2({\tilde{x}}(p_2), p_2)$ using Theorem~\ref{lemma:Lemma:4}. Since from (\ref{eqn:114}), we know that 
$f_1(x,p_2) > f_2(x,p_2)  \,\forall \, x \in [x_1^{*}(p_2),{\tilde{x}}(p_1) ]$, we  conclude that,

$$ {\tilde{x}}(p_2) \not \in [x_1^{*}(p_2), \tilde{x}(p_1)] \textrm{ which implies  } {\tilde{x}}(p_2) \geq {\tilde{x}(p_1)}.$$

\bigskip\

Under $A_{12}$, using Proposition~\ref{lemma:lemma101}, we get $f_2(x_1^{*}(p_2), p_2) \geq f_1(x_1^{*}(p_2), p_2)$ which violates (\ref{eqn:114}). Hence,  $x_1^*(p_2)< \tilde{x}(p_1)$ cannot hold.

\bigskip

{\bf Case $A_{11}$ :} 

 \bigskip

Under \textbf{(C1)} for $p= p_1$ and $p=p_2$, using Proposition \ref{lemma:lemma101}, we get ${\tilde{x}}(p_2) = x_2^{*}(p_2)$ and ${\tilde{x}}(p_1) = x_2^{*}(p_1)$.

Using Lemma \ref{lemma:1}, we get $x_2^{*}(p_2) \geq x^{*}_2(p_1)$ hence,
$${\tilde{x}}(p_2) \geq {\tilde{x}(p_1)}.$$

\vskip 0.2in

\section{Supporting Lemmas}\label{app:lemma}
We now state and prove the lemmas that are used in proving the the results of the paper in the previous appendix.

\begin{lemma}\label{lemma:badset}
	For $\epsilon_1>0$ and $\epsilon_2>0$ in the neighbourhood of zero,  we have 
	$$
	\mathbb{P}(\mathcal{G}_t^c)\le 4\left[\exp\left(-\frac{tc_1^2\epsilon_1^2}{2\sigma}\right)+ \exp\left(-\frac{t\epsilon_2^2}{8}\right)+t\exp\left(-\frac{t\epsilon_2 c_1^2\eta^2}{6\sigma}\right) \right],
	$$
	where $c_1$, $\sigma$, $\gamma$ and $\eta$ are positive constants.
\end{lemma}

	We define following terms which will be used in the proof.
	\begin{eqnarray}\label{martingale}
	C_t=\sum_{j=1}^t\left(J_j-\tilde{w}(\hat{p}_{j-1})\right).\\
	D_t = \sum_{j=1}^t\left( \Psi (I_j,X_j,p) - \Psi(I_j,X_j,\hat{p_t})\right).
	\end{eqnarray}
	
	\begin{equation}\label{psi}
	\Psi (I_j,X_j,p)= \left(\frac{I_{j} \,h^{'}(X_j,p) }{h(X_j,p)(1-h(X_j,p))}
	-\frac{h^{'}(X_j,p)}{1-h(X_j,p)} \right).
	\end{equation}

We also state a supporting lemma which will be used in the proof and it will be proved later.

\begin{lemma}\label{lipchitz}
	If Assumpution 3.2  is satisfied by $h(x,p)$, $p \in [\underline{p}, \bar{p}]$ and $x \in [\underline{x}, \bar{x}]$ then,

	\begin{equation}\label{reverselipchitz}
	c_2|\hat{p}_t -p| \ge| \left( \Psi (I_j,X_j,p) -  \Psi(I_j,X_j,\hat{p_t})\right)|  \ge c_1|\hat{p}_t -p|,
	\end{equation}
	
	where $c_1$ are $c_2$  are positive constants.
\end{lemma}

	\bigskip{}
\noindent\textbf{Proof of Lemma \ref{lemma:badset}:}

	First we show that $C_t$ and $D_t$ are martingales.  
	Since $E(J_j) = \tilde{w}(\hat{p}_{j-1})$, $J_j$ is an indicator function and $\tilde{w}(\hat{p}_{j-1}) \in [0,1]$. Hence $C_t$ is a martingale with bounded increments (bounded by 1). To see that $D_t$ is a martingale, first observe,
	
\begin{equation}\label{temp1}
    E\left(\frac{D_{t+1}}{D_t}\right) =  D_t + E\left( \Psi (I_{t+1},X_{t+1},p) - \Psi(I_{t+1},X_{t+1},\hat{p_t})\right
	).
\end{equation}
	From \eqref{eqn:estiamte_p} we know that $\Psi(I_{t+1},X_{t+1},\hat{p_t}) = 0$. Hence on substituting the value of $\psi(I_{t+1}, X_j, p)$ from \eqref{psi} in \eqref{temp1} we get,
	
	$$E\left(\frac{D_{t+1}}{D_t}\right) =  D_t + E \left(\frac{I_{t+1} \,h^{'}(X_{t+1},p) }{h(X_{t+1},p)(1-h(X_{t+1},p))}
	-\frac{h^{'}(X_{t+1},p)}{1-h(X_{t+1},p)} \right).$$
	
	Since $E(I_{t+1}) = h(X_{t+1}, p)$, hence,
	
		$$E\left(\frac{D_{t+1}}{D_t}\right) =  D_t. $$
	
		\bigskip
	
Let  $ A \triangleq \{|\hat{p}_t-p| \ge \epsilon_1\}$ and $R \triangleq \left\{\left|\frac{1}{t}\sum_{j=1}^{t}(J_j-\tilde{w}(p))\right|\ge \epsilon_2\right\}$.

\bigskip

Observe that  $ \mathbb{P}(\mathcal{G}_t^c) \leq \mathbb{P}(A) + \mathbb{P}(R)$.  We upper bound  $ \mathbb{P}(A)$ as well as $\mathbb{P}(R)$.
	
	\bigskip
	
First consider $\mathbb{P}(A)$. Let  $E_1 = \{\hat{p_t} \in [\frac{\underline{p}}{\gamma},\gamma \bar{p}] \}$,  $E_2 = \{\hat{p_t} \in (\gamma \bar{p},\infty) \}$ and  $E_3 = \{\hat{p_t} \in (-\infty,\frac{\underline{p}}{\gamma}) \}$, where $\gamma$ is a large positive constant. Since true value of $p \in (\underline{p},\bar{p})$, both  $E_2$ and $E_3$ are rare events.
	
	\bigskip{}
	
	Observe that,
	$$\mathbb{P}(A) \le \mathbb{P}(A \cap E_1) + \mathbb{P}(E_2)+  \mathbb{P}(E_3). $$

	\textbf{Bounding $\mathbb{P}(A \cap E_1)$ from above:}
	
	\bigskip
To see that $D_t$ is a martingale with bounded increments under the event $A\cap E_1$, observe that,    
$$D_t = \sum_{j=1}^t\left( \Psi (I_j,X_j,p) - \Psi(I_j,X_j,\hat{p_t})\right).$$

Using Assumption 3.2 , from Lemma \ref{lipchitz}, we get

$$	c_2|\hat{p}_t -p| \ge| \left( \Psi (I_j,X_j,p) -  \Psi(I_j,X_j,\hat{p_t})\right)| .$$
It follows that,

$$|D_{t+1}- D_t| \leq c_2|\hat{p}_t -p|.$$
Since $p \in (\underline{p}, \bar{p})$ and under the event $E_1$, $\hat{p_t} \in [\frac{\underline{p}}{\gamma},\gamma \bar{p}]$, hence we get,

$$|D_{t+1}- D_t| \leq \sigma,
$$
 where $\sigma = c_2 \textrm{max} \{(\gamma \bar{p} - \underline{p} ), (\bar{p}- \frac{\underline{p}}{\gamma}) \}$.

It follows that $D_t$ is a martingale with bounded increments (bounded by $\sigma$). Using Azuma-Hoeffding on $D_t$, we get:
	$$
	\mathbb{P}\left(\sum_{j=1}^t\left( \Psi (I_j,X_j,p) - \Psi(I_j,X_j,\hat{p_t})\right)>t\theta\right)\le 2 \exp\left(-\frac{t^2\theta^2}{2t\sigma}\right)=2 \exp\left(-\frac{t\theta^2}{2\sigma}\right),
	$$
	for all $\theta>0$.
	Using Assumption 3.2 under event $E_1$, from Lemma \ref{lipchitz}, we conclude that
	$$
	\frac{1}{t}\sum_{j=1}^t\left( \Psi (I_j,X_j,p) - \Psi(I_j,X_j,\hat{p_t})\right)\ge c_1|\hat{p}_t-p|.
	$$
	Thus, 
	\begin{equation}
	\mathbb{P}(A \cap E_1)\le \mathbb{P}\left(\frac{1}{t}\sum_{j=1}^t\left( \Psi (I_j,X_j,p) - \Psi(I_j,X_j,\hat{p_t})\right)>c_1\epsilon_1\right)\le 2 \exp\left(-\frac{tc_1^2\epsilon_1^2}{2\sigma}\right).
	\label{eqn:bound1}
	\end{equation}
	
	\bigskip

	\textbf{Bounding $\mathbb{P}(E_2)$ from above:}
	
	\bigskip
	
	Recall that $\hat{p_t}$ solves (\ref{eqn:estiamte_p}). Suppose if we had asked the hardest question i.e. $\bar{x}$ and kept the responses of the candidate, i.e., $(I_j )$ to be same and then we get the solution of (\ref{eqn:estiamte_p}) to be $\bar{p_t}$ that satisfies the following equation,
	
	\begin{equation}\label{mlle}
	\sum_{j=1}^{t}\frac{I_{j} \,h^{'}(\bar{x},\bar{p_t}) }{h(\bar{x},\bar{p_t})(1-h(\bar{x},\bar{p_t}))}
	=\sum_{j=1}^{t}\frac{h^{'}(\bar{x},\bar{p_t})}{1-h(\bar{x},\bar{p_t})}\,\, \forall I_j = {1,0}.    
	\end{equation}
	
Now further observe that (\ref{mlle}) simplifies to,	
	\begin{equation}\label{pbart}
	\sum_{j=1}^{t}\frac{I_{j}  }{t}
	= h(\bar{x},\bar{p_t} ).   
	\end{equation}

	Under Assumption 3.2, we get the single crossing property of functions $\log h(x,p)$ and $\log (1-h(x,p))$. Using that one can see that in (\ref{eqn:estiamte_p}), keeping all $(I_j \textrm{ for } j = 1 \textrm{ to } t)$ fixed and increasing any $X_j$ results in increase of $\hat{p_t}$. Hence we conclude that $\bar{p_t}> \hat{p_t}$.
	
	\bigskip
	Since $\bar{p_t} > \hat{p_t}$,  using (\ref{pbart}) we get,
	$$\mathbb{P}(E_2) \le \mathbb{P} (\bar{p_t} > \gamma \bar{p})=\mathbb{P} \left( \sum_{j=1}^{t}\frac{I_{j}  }{t}  > h(\bar{x}, \gamma \bar{p})\right) . $$
	
	Suppose if we ask the easiest question i.e. $\underline{x}$ to the candidate with the highest ability ,i.e., $p = \bar{p}$ and the responses to the questions asked are given by $\tilde{I_j}$ then by simple coupling argument we get,
	
	$$ \mathbb{P} \left( \sum_{j=1}^{t}\frac{I_{j}  }{t}  > h(\bar{x}, \gamma \bar{p})\right)  \leq \mathbb{P}\left( \sum_{j=1}^{t}\frac{\tilde{I_{j}}  }{t}  > h(\bar{x}, \gamma \bar{p})\right).$$
	Hence,
	
	$$\mathbb{P}(E_2) \le \mathbb{P} \left( \sum_{j=1}^{t}\frac{\tilde{I_{j}}  }{t} - h(\underline{x}, \bar{p}) > h(\bar{x}, \gamma \bar{p})- h(\underline{x}, \bar{p})\right).$$
	
	Using Hoeffding inequality,
	
	\begin{equation}\label{E2}
	\mathbb{P}(E_2) \le  \exp\left(-2t( h(\bar{x}, \gamma \bar{p})- h(\underline{x}, \bar{p}))^2\right).
	\end{equation}

	Similarly, we get the upper bound on $\mathbb{P}(E_3)$, i.e.,
	
	\begin{equation}\label{E3}
	\mathbb{P}(E_3) \le  \exp\left(-2t( h(\underline{x}, \frac{\bar{p}}{\gamma})- h(\bar{x}, \underline{p}))^2\right).
	\end{equation}
	
	Combining (\ref{eqn:bound1}), (\ref{E2}) and (\ref{E3}), we have,
	
	\[
	\mathbb{P}(A) \le \exp\left(-2t( h(\bar{x}, \gamma \bar{p})- h(\underline{x}, \bar{p}))^2\right)+ \exp\left(-2t( h(\underline{x}, \frac{\bar{p}}{\gamma})- h(\bar{x}, \underline{p}))^2\right)+ 2 \exp\left(-\frac{tc_1^2\epsilon_1^2}{2\sigma}\right).
	\]

	For $\epsilon_1$ sufficiently small, we can choose $\gamma$ such that $( h(\bar{x}, \gamma \bar{p})- h(\underline{x}, \bar{p}))$ and $( h(\underline{x}, \frac{\bar{p}}{\gamma})- h(\bar{x}, \underline{p}))$ are bounded below by $\epsilon_1$. Hence,
	
	\begin{equation}\label{PA}
	\mathbb{P}(A)  =\mathbb{P} \{|\hat{p}_t-p| \ge \epsilon_1\}  \le  4 \exp\left(-\frac{tc_1^2\epsilon_1^2}{2\sigma}\right).
	\end{equation}

	\bigskip
	
	\textbf{Bounding $\mathbb{P}(R)$ from above :}

	\begin{eqnarray*}
		\mathbb{P}\left(\left|\frac{1}{t}\sum_{j=1}^{t}(J_j-\tilde{w}(p))\right|\ge \epsilon_2\right)&=&\mathbb{P}\left(\left|\frac{1}{t}\sum_{j=1}^{t}(J_j-\tilde{w}(\hat{p}_{j-1}))+(\tilde{w}(\hat{p}_{j-1})-\tilde{w}(p))\right|\ge \epsilon_
		2\right)\\
		&\le&\mathbb{P}\left(\left|\frac{1}{t}\sum_{j=1}^{t}(J_j-\tilde{w}(\hat{p}_{i-1}))\right|\ge \frac{\epsilon_2}{2}\right)+\mathbb{P}\left(\left|\frac{1}{t}\sum_{j=1}^{t}(\tilde{w}(\hat{p}_{i-1})-\tilde{w}(p))\right|\ge \frac{\epsilon_2}{2}\right).
	\end{eqnarray*}
	For the first term, we can use the fact that $C_t$ is a martingale with bounded increments (bounded by 1) and thus using Azuma-Hoeffding we obtain 
	\begin{equation}\label{eqn:bound2}
	\mathbb{P}\left(\left|\frac{1}{t}\sum_{j=1}^{t}(J_j-\tilde{w}(\hat{p}_{i-1}))\right|\ge \frac{\epsilon_2}{2}\right)\le 2\exp\left( -\frac{t\epsilon_2^2}{8}\right).
	\end{equation}
	For the second term, note that 
	\begin{eqnarray*}
		\mathbb{P}\left(\left|\frac{1}{t}\sum_{j=1}^{t}(\tilde{w}(\hat{p}_{j-1})-\tilde{w}(p))\right|\ge \frac{\epsilon_2}{2}\right)&\le& \mathbb{P}\left(\frac{1}{t}\left(t_0+\sum_{j=t_0}^{t}|\tilde{w}(\hat{p}_{j-1})-\tilde{w}(p)|\right)\ge \frac{\epsilon_2}{2}\right)\\
		&=& \mathbb{P}\left(\left(\sum_{j=t_0}^{t}|\tilde{w}(\hat{p}_{j-1})-\tilde{w}(p)|\right)\ge \frac{t\epsilon_2}{2}-t_0\right),\\
		&\le& \sum_{j=t_0}^{t}\mathbb{P}\left(|\tilde{w}(\hat{p}_{j-1})-\tilde{w}(p)|\ge \frac{\epsilon_2}{2}-\frac{t_0}{t}\right).
	\end{eqnarray*}
	where $t_0<t$. For $\epsilon_2$ small, we can choose $t_0=\frac{\epsilon_2 t}{3}$, we then have the above probability is bounded by 
	$$
	\sum_{j=t_0}^{t}\mathbb{P}\left(|\tilde{w}(\hat{p}_{j-1})-\tilde{w}(p)|\ge \frac{\epsilon_2}{6}\right).
	$$
	
	Note that using the continuity of $\tilde{w}(\cdot)$, which is true because of maximum theorem of optimization, we have that there exists $\eta>0$ such that if $|q-p|<\eta$ then $|\tilde{w}(q)-\tilde{w}(p)|<\frac{\epsilon_2}{6}$. Combining with (\ref{PA}),
	\begin{eqnarray}\label{eqn:bound3}
	\mathbb{P}\left(|\tilde{w}(\hat{p}_{j-1})-\tilde{w}(p)|\ge \frac{\epsilon_2}{6}\right)
	\le \mathbb{P}(|\hat{p}_j-p| \geq \eta)\le 4\exp\left(-\frac{jc_1^2\eta^2}{2\sigma}\right),
	\end{eqnarray}
	for all $j\ge t_0$.
	
	Combining bounds in equations (\ref{PA}), (\ref{eqn:bound2}) and (\ref{eqn:bound3}), we obtain our result.

\bigskip

\textbf{Proof of the lemma~\ref{lemma:dec_fun}.}

$$ \textrm{We can re-express } B(x) \textrm{ as }\frac{I(x,u_{i},p)}{I(x,u_{i+1},p)}.$$

$$ \textrm{Where, }I(x,u_{i}) = {a(p,u_{i})x + b(p,u_{i})}, \,\, a(p,u_{i}) = \left(p-u_{i} + p\log\left(\frac{u_{i}}{p}\right)\right)\textrm{ and  } b(p,u_{i})= \left(p^2 + pu_{i}\left(\log\left(\frac{u_{i}}{p}\right)-1\right)\right).$$

We show that $B(x)$ is a strictly decreasing function for $x$  $\in$ $(x_1^*, x_2^*  )$ by exploiting the properties of $I(x,u_{i},p)$ and $I(x,u_{i+1},p)$. Hence we first analyse $I(x,u_{i},p)$ and $I(x,u_{i+1},p)$.
\bigskip

Observe that,
$$ \frac{\partial I(x,u_{i},p)}{\partial x} =a(p,u_{i})=  p\left( \log \left(\frac{u_{i}}{p}\right)- \left(\left(\frac{u_{i}}{p}\right)-1\right)\right).$$
Since $\log x < (x-1) \, \textrm{for all} \,x >0 \,\, \textrm{and}\, x\neq1,$

\begin{equation}{\label{sign_I}}
\frac{\partial I(x,u_{i},p)}{\partial x} = a(p,u_{i})<0, \,\, \forall u_{i},\,p >0 \textrm{ and }u_{i} \neq p.  
\end{equation}

It follows that  $I(x,u_{i},p)$  is strictly decreasing in $x$ for $x$  $\in$ $(x_1^*, x_2^*  )$. Similarly we can prove that $I(x,u_{i+1},p)$ is also strictly decreasing in $x$ for $x$  $\in$ $(x_1^*, x_2^*  )$.
\bigskip

We know from (\ref{eqn:max_x}) that root of $I(x,u_{i},p)= 0$  is $x_1^{*}$ since it is the root of $f_1^{'}(x)$ for $h(x,p) = \frac{p}{p+x}$ from its definition. Similarly the root of $I(x,u_{i+1},p) = 0$ is  $x_2^{*}$.
\begin{equation}{\label{sign_X}}
x^*_1 =- \frac{b(p,u_{i})}{a(p,u_{i})} \textrm{ and }x^*_2 = -\frac{b(p,u_{i+1})}{a(p,u_{i+1})}.
\end{equation}

Since $I(x,u_{i},p)$ and $I(x,u_{i+1},p)$ are  strictly decreasing functions of $x$ hence for $x$  $\in$ $(x_1^*,  x_2^*)$, 

\begin{equation}{\label{sign_B(x)}}
I(x,u_{i},p)\leq 0 \,\,\textrm{and},\,\,I(x,u_{i+1},p)\geq 0.    
\end{equation}

From (\ref{sign_B(x)}) we conclude that,
$$B(x)\geq{0} \,\, \forall x \in (x_1^*,  x_2^*).$$

\bigskip

The derivative of the function $B(x)$ w.r.t. $x$ is given by,
$$\frac{d B(x) }{d x} = \frac{\frac{\partial I(x,u_{i},p)}{\partial x}I(x,u_{i+1},p) - \frac{\partial I(x,u_{i+1},p)}{\partial x}I(x,u_{i},p)}{(I(x,u_{i+1},p))^{2}}.$$ 

Using (\ref{sign_I}) and (\ref{sign_B(x)}) we get,
$$ B^{'}(x)< 0\,\, \forall \,\, x  \in (x_1^*,  x_2^*).$$

\bigskip

Observe that, 
\begin{equation}{\label{LbyL'}}
\frac{B(x)}{B{'}(x)} = \frac{(a(p,u_{i})x+b(p,u_{i}))(a(p,u_{i+1})x+b(p,u_{i+1}))}{a(p,u_{i})b(p,u_{i+1})-b(p,u_{i})a(p,u_{i+1})}.
\end{equation}

Hence,

$$ \frac{d^{2}[\frac{B(x)}{B{'}(x)}]}{dx^{2}}= \frac{2a(p,u_{i})a(p,u_{i+1})}{a(p,u_{i})b(p,u_{i+1})-b(p,u_{i})a(p,u_{i+1})} .$$

We want to show that above expression is negative. To prove that first we observe that $b(p,u_{i})>0$  $\forall \,u_{i},\,p >0 \textrm{ and }u_{i}\neq p$ since  $x\log x >{(x-1)}$  $\forall  \,x>0 \textrm{ and }x\neq 1$. Similarly $b(p,u_{i+1}) > 0$ $\forall \,u_{i+1},\,p >0 \textrm{ and }u_{i+1}\neq p$. Since $x^*_1 < x^*_2$ and combining it with (\ref{sign_X}) we conclude that $a(p,u_{i})b(p,u_{i+1})-b(p,u_{i})a(p,u_{i+1}) <0$.

From (\ref{sign_I}), we get that $a(p,u_{i})<0$ and  $a(p,u_{i+1})<0$   $\forall \,u_{i},\,u_{i+1},\,p >0 \textrm{ and }u_{i},\,u_{i+1} \neq p$. Hence combining these results we can conclude that $\frac{d^{2}[\frac{B(x)}{B{'}(x)}]}{dx^{2}}< 0$ which concludes our proof.

\bigskip

\noindent\textbf{Proof  of Lemma \ref{lipchitz}:}
	
	Assumption 3.2 implies that lower bounds in the (\ref{derbound}) and upper bounds in the (\ref{derbound}) come from twice continuous differentiablity of $h(x,p)$ when $x$ and $p$ lies in a compact interval.

	\begin{equation}\label{derbound}
	k_2\ge|\frac{\partial \log(h(x,p))}{\partial p^2}| \ge  k_1, \,\,   k_4\ge|\frac{\partial \log(1-h(x,p))}{\partial p^2}| \ge  k_3.
	\end{equation} 
	
	Where  $k_1,\,\, k_2, \,\,k_3$ and $k_4$ are positive constants.

	\bigskip
	
	Equation (\ref{derbound}) can be re-expressed as,

	\begin{equation}\label{derivative1}
	k_2|p_1-p_2|\ge \left|\frac{\partial  \log h(x_1,p)}{\partial p} \vert_{p = p_1} -\frac{\partial  \log h(x_2,p)}{\partial p}\vert_{p = p_2} \right |\ge  k_1|p_1-p_2|,   
	\end{equation}

	\begin{equation}\label{derivative2}
	k_4|p_1-p_2|\ge \left|\frac{\partial  \log (1-h(x_1,p))}{\partial p} \vert_{p = p_1} -\frac{\partial  \log (1-h(x_2,p))}{\partial p}\vert_{p = p_2} \right |\ge  k_3 |p_1-p_2|,  
	\end{equation}
	
	$$\forall p_1, p_2 \in [\underline{p}, \bar{p}] \textrm{ and }  \forall x_1, x_2 \in [\underline{x}, \bar{x}].$$

	\bigskip
	
	Observe that for $I_j = 0$,
	
	$$ |\left( \Psi (0,X_j,p) -  \Psi(0,X_j,\hat{p_t})\right)| =\left|\frac{\frac{\partial h(X_j,p)}{\partial p}}{(1-h(X_j,p) )} - \frac{\frac{\partial h(X_j,\hat{p}_t)}{\partial \hat{p}_t}}{(1-h(X_j,\hat{p}_t) )}\right|,$$
	
	and for $I_j = 1$, we get,
	
	$$|\left( \Psi (1,X_j,p) -  \Psi(1,X_j,\hat{p_t})\right)| =\left|\frac{\frac{\partial h(X_j,p)}{\partial p}}{(h(X_j,p) )} - \frac{\frac{\partial h(X_j,\hat{p}_t)}{\partial \hat{p}_t}}{(h(X_j,\hat{p}_t) )}\right|.$$
	
	One can see from (\ref{derivative1}) and (\ref{derivative2})  that we get tht \eqref{lipchitz} follows  for both cases i.e. when $I_j$ takes value zero or one.

\bigskip

\vskip 0.2in


\begin{lemma} \label{uniquenessofmle}
	For  $f(x)$  with $h(x,p)$ of the form (\ref{eqn:1}), Assumption~ 3.1 holds.
\end{lemma}

\noindent\textbf{Proof :}
	Note that the likelihood of observing data $(I_j: 1 \leq j \leq t)$ when the underlying ability is $p$ and the questions are asked at level ${\bf X}_t$ 
	is given by
	\[
	L({\bf X}_t, p) = \prod_{j=1}^t \left ( \frac{g(p)}{g(p)+k(X_j)}\right )^{I_j} \left ( \frac{k(X_j)}{g(p)+k(X_j)}\right )^{1-I_j}
	\]
	and the  log-likelihood equals
	\[
	\log L({\bf X}_t, p) = \sum_{j=1}^t I_j \log \left(\frac{g(p)}{g(p)+k(X_j)}\right) + (1-I_j) \log \left(\frac{k(X_j)}{g(p)+k(X_j)}\right).
	\]
	
	It follows that,
	
	$$\frac{\partial \log L(p, X_t)}{\partial p} = \sum_{j=1}^t \left[ I_j - \left( \frac{g(p)}{g(p) + k(X_j)} \right) \right] .$$
	
	Hence MLE solution $\hat{p}_t$ satisfies,
	
	$$\sum_{j=1}^t  I_j = \sum_{j=1}^t \left( \frac{g(\hat{p}_t)}{g(\hat{p}_t) + k(X_j)} \right) .$$
	It easily follows that for $p< \hat{p}_t$,  $\frac{\partial L(p, X_t)}{\partial p} >0$ and for $p> \hat{p}_t$, $\frac{\partial L(p, X_t)}{\partial p } <0$  . This concludes the proof.

\bigskip

\begin{lemma} \label{supermod}
	For  $f(x)$  with $h(x,p)$ of the form (\ref{eqn:1}), Assumption~ 3.2 holds.
\end{lemma}

\noindent\textbf{Proof :}
	\textbf{Part (a)} This result directly follows from the definition.
	
	\bigskip
	\textbf{Part (b)} For $h(x,p) = \frac{g(p)}{g(p)+k(x)}$, we get 
	
	$$ \frac{\partial \log(h(x,p))}{\partial x\partial p} = \frac{\partial \log(1-h(x,p))}{\partial x\partial p}= \frac{g^{'}(p)k^{'}(x)}{(g(p)+k(x))^2} \ge 0$$
	
	Last inequality follows from  strictly increasing property of
	  $g(.)$ and $k(.)$.

\begin{lemma} \label{lemma:1}
	For  $f(x)$  with $h(x,p)$ of the form (\ref{eqn:1}), $p \ne u$, 
	Assumption~\ref{ass1} holds. Furthermore,  the point where $f(x)$ achieves its maximum value, i.e., $x^*(p,u)$,  is non decreasing w.r.t. both  $p$ and $u$.
\end{lemma}

\noindent\textbf{Proof :}
	
	Observe that  for $h(x,p) = \frac{g(p)}{g(p)+k(x)}$,  $f(x)= d(h(x,p)| h(x,u))$ equals
	\[
	\log \frac {\tilde{u}+k(x)}{\tilde{p}+k(x)}  -  \frac{\tilde{p}}{\tilde{p}+k(x)} \log \frac{\tilde{u}}{\tilde{p}}.
	\]
	Where $\tilde{p} = g(p)$ and $\tilde{u}= g(u)$, since $g(\cdot)$ is a strictly increasing function hence for each $p$ and $u$, we get  unique $\tilde{p}$ and $\tilde{u}$ respectively.

	Setting $y=k(x)/\tilde{u}$ and $z = \tilde{u}/\tilde{p}$, $f(u y)$ quals
	\[
	\log z +\log (1+y) - \log (1+z y) - \frac{1}{1+zy} \log z.
	\]
	Differentiating  w.r.t. $y$, and after some simplifications, 
	\begin{equation} \label{eqn:701}
	\frac{df(uy)}{dy} = \frac{(z \log z -z+1) + y z(\log z +1 -z)}{(1+y)(1+wy)^2}.
	\end{equation}

	One can see that denominator in R.H.S of (\ref{eqn:701}) is always positive and numerator is linear in $y$ hence monotone in $x$ since $k(\cdot)$ is an increasing function. Thus,  the derivative uniquely equals zero at $x^*(p,u)$ that satisfies,
	
	$$k(x^{*}(p,u)) =\tilde{p}\frac{z \log z -(z-1)}{(z-1)- \log z}.$$
	
	It is easy to see that both numerator and denominator are positive for $z>0$ and $z \neq 1$ in the expression of $x^*(p,u)$. Hence,
	\begin{equation} \label{eqn:max_x}
	x^{*}(p,u) = k^{-1}\left(\frac{\tilde{p}\tilde{u}- {\tilde{p}}^2 - \tilde{p}\tilde{u} \log\frac{\tilde{u}}{\tilde{p}}}{\tilde{p}-\tilde{u} +\tilde{p}\log\frac{\tilde{u}}{\tilde{p}}}\right).
	\end{equation}
	
	One can also see from (\ref{eqn:701}) that  the derivative of $f(x)$ is greater than zero for  $x<k^{-1}(x^*(p,u))$ and less than zero for $x> k^{-1}(x^*(p,u))$.
	Thus, $f(x)$ is a quasi concave function of $x$.

	\bigskip{}

	\textbf{ Non-decreasing property of $x^*(p,u)$ w.r.t. $p$ and $u$.}
	
	From (\ref{eqn:max_x}), we know that $x^*(p,u)$ for  $h(x,p) = \frac{g(p)}{g(p)+k(x)}$ is $k^{-1}(\frac{\tilde{p}\tilde{u}- {\tilde{p}}^2 - \tilde{p}\tilde{u} \log\frac{\tilde{u}}{\tilde{p}}}{\tilde{p}-\tilde{p} +\tilde{p}\log\frac{\tilde{u}}{\tilde{p}}})$, and we know that $k^{-1}$ is an increasing function and $\tilde{p}$,  $\tilde{u}$ are increasing functions in p and u respectively. Hence it suffices to prove the non decreasing property of 
	$x^*(p,u)$ when  $h(x,p) = \frac{p}{p+x}$.
	
	Hence to prove the non-decreasing property of $x^*(p,u)$ for $h(x,p) = \frac{g(p)}{g(p)+k(x)}$, we prove it for $h(x,p) = \frac{p}{p+x}$.

	\bigskip

	Set $t = \frac{u}{p}$ and observe that,
	
	$$\frac{\partial x_1^{*}(p,u)}{\partial p} = \frac{(t-1)(2(1-t) + (1+t)\log(t)) }{p(\log(t) + 1-t)^2}.$$
	
	Observe that the denominator is always positive for $p, u>0$. To see that numerator is always non negative. Let,
	$$N(t) =  (t-1)(2(1-t) + (1+t)\log(t)).$$

	First we prove that it is a convex function for $t>0$, then we show that it's minimum value is zero hence $N(t)$ and $\frac{\partial x_1^{*}(p,u)}{\partial p}$  is always non-negative.
	
	Observe that,
	$$N^{'}(t) = 4 - 3t + 2t \log(t) -\frac{1}{t}, \textrm{ and } N^{''}(t) = \frac{t^{2}\log(t^2) - (t^2 -1)}{t^2}.$$
	
	It is easy to see that $N^{''}(t) \geq{0}$ since $x\log x \geq{(x-1)}$. Hence $N(t)$ is a convex function. Further observe $N^{'}(1) = 0$ hence t = 1  is a local extrema and because of convexity, $t =1$ will be the global minima.
	
	Minimum value of $N(t)$ will be achieved at t =1 which is $0$ hence this completes the proof of the claim.

	\bigskip
	To observe that $x^{*}(p,u)$ is non decreasing w.r.t. u, set $t = \frac{u}{p}$,
	
	$$\frac{\partial x_1^{*}(p,u)}{\partial u} = \frac{(1-t)^2 - t{(\log(t))}^2}{t(\log(t) + 1-t)^2}.$$
	
	Observe that denominator is always positive. Let $\tilde{N}(t)$  be  the numerator by .
	$$\tilde{N}(t) = (t-1)^2 - t{(\log(t))}^2.$$
	To see that $\tilde{N}(t)$ is always non-negative, observe that,
	$$\Tilde{N}^{'}(t) = 2(t-1) - {(\log(t))}^2 - 2\log(t) \textrm{, and } \tilde{N}^{''}(t) = 2\left( \frac{(t-1) - \log(t)}{t} \right).$$

	One can easily observe that $\tilde{N}^{''}(t)$ is non negative for $t>0$ which is our domain hence $\tilde{N}(t)$ is convex for $t>0$. We can easily check at $t =1$, $\tilde{N}^{'}(t) =0 $ hence as similar to previous argument $t =1$ is the global minima of $\tilde{N}(t)$. 
	
	\bigskip
	
	$\tilde{N}(1) = 0$ implies $\tilde{N}(t)$ is always non- negative and hence $\frac{\partial x_1^{*}(p,u)}{\partial u}$ is also non- negative which completes the proof.
	$\Box$

\section{Algorithm to solve dual problem when $\cal{X}$ is discrete. }\label{app:algo}

Here, we outline the procedure to compute the solution to 
$\max \,(y_1+y_2)$, such that $a_{i} y_1 + b_{i} y_2 \leq 1$ for $i = 1, \ldots, k$, and $y_1, y_2 \geq 0$, in $O(k)$ time,
when the vectors $(a_{i}, b_{i})$ are strictly monotone in the  second argument, that is $b_{1} >
b_{2} > \ldots > b_{k}$.

To see this consider the positive quadrant corresponding to $y_1 \geq 0$ and $y_2 \geq 0$.   We arrive at the lower envelope, restricted to the positive quadrant,
of the lines associated with the $k$ constraints when they are tight. 
The outline of determining this lower envelope is as follows:
\begin{enumerate}
	\item
	First line gives the lower envelope as the line segment between $(0,b_{1}^{-1})$ and
	$(a_{1}^{-1},0)$. 
	\item
	Suppose after line $m (1 \leq m < k)$ has been considered, the running lower envelope is denoted by the segments obtained by sequentially joining  points
	\[
	(\tilde{a}_1, \tilde{b}_1),  \ldots, (\tilde{a}_r, \tilde{b}_r)
	\]
	for $1 \leq r \leq m$, and $(\tilde{a}_1, \tilde{b_1})= (0,b_{1}^{-1})$.  Furthermore, the adjacent points correspond 
	to segments in one of the $m$ lines already considered.
	\item
	When the segment joining $(0, b_{m+1}^{-1})$ to $ (a_{m+1}^{-1},0)$  in the positive quadrant corresponding to line $m+1$ is considered, if its slope is greater than the slope of the last segment in the running lower envelope,  $((\tilde{a}_{r-1}, \tilde{b}_{r-1}),(\tilde{a}_r, \tilde{b}_r))$, i.e., 
	\[
	\frac{b_{m+1}^{-1}}{-a_{m+1}^{-1}} > \frac{\tilde{b}_r- \tilde{b}_{r-1}}{\tilde{a}_r-\tilde{a}_{r-1}},
	\]
	then the line $m+1$ is fathomed and we move on to the next line. 
	\item
	Else, find $s \leq r$ such that 
	\[
	\frac{\tilde{b}_{s-1}- \tilde{b}_{s-2}}{\tilde{a}_{s-1}-\tilde{a}_{s-2}} < \frac{b_{m+1}^{-1}}{-a_{m+1}^{-1}} <  \frac{\tilde{b}_s- \tilde{b}_{s-1}}{\tilde{a}_s-\tilde{a}_{s-1}}.
	\]
	If $s <k$, the segments  corresponding to sequentially joining points $(\tilde{a}_{s+1}, \tilde{b}_{s+2}), \ldots, (\tilde{a}_r, \tilde{b}_r))$  are no longer in the lower envelope and are fathomed. 
	\item
	The updated lower envelope consists  of
	\[
	\left ( (\tilde{a}_1, \tilde{b}_1),  \ldots, (\tilde{a}_{s-1}, \tilde{b}_{s-1}), (\tilde{a}_{s^*}, \tilde{b}_{s^*}),
	(\tilde{a}_{m+1}, \tilde{b}_{m+1})\right)
	\]
	where $(\tilde{a}_{s^*}, \tilde{b}_{s^*})$ denotes the point of intersection between line
	$(0, b_{m+1}^{-1})$ to $ (a_{m+1}^{-1},0)$,  and the line segment joining 
	$(\tilde{a}_{s-1}, \tilde{b}_{s-1})$ to $(\tilde{a}_{s}, \tilde{b}_{s})$. Furthermore,
	$(\tilde{a}_{m+1}, \tilde{b}_{m+1})= (a_{m+1}^{-1},0)$.
\end{enumerate}

The algorithm has $k$ steps, at each step $m+1$, if Step 4 is reached and $s<k$,
at least one line segment corresponding one of the $k$ lines is fathomed, and each segment maybe be  fathomed at most once.
If Step 4 is reached and $s=k$, $O(1)$ computation is performed. Hence,  the algorithm running time is $O(k)$.

Once the lower envelope is constructed, a simple sweep across can be conducted in $O(k)$ time to find a point that maximizes
$y_1+y_2$.
\newpage

\section{Figure for inbulit exploration numerical study}\label{s:study}

\begin{figure}[h!] \label{easy}\label{normal}\label{hard}
\begin{center}
\includegraphics[height=0.4\textheight]{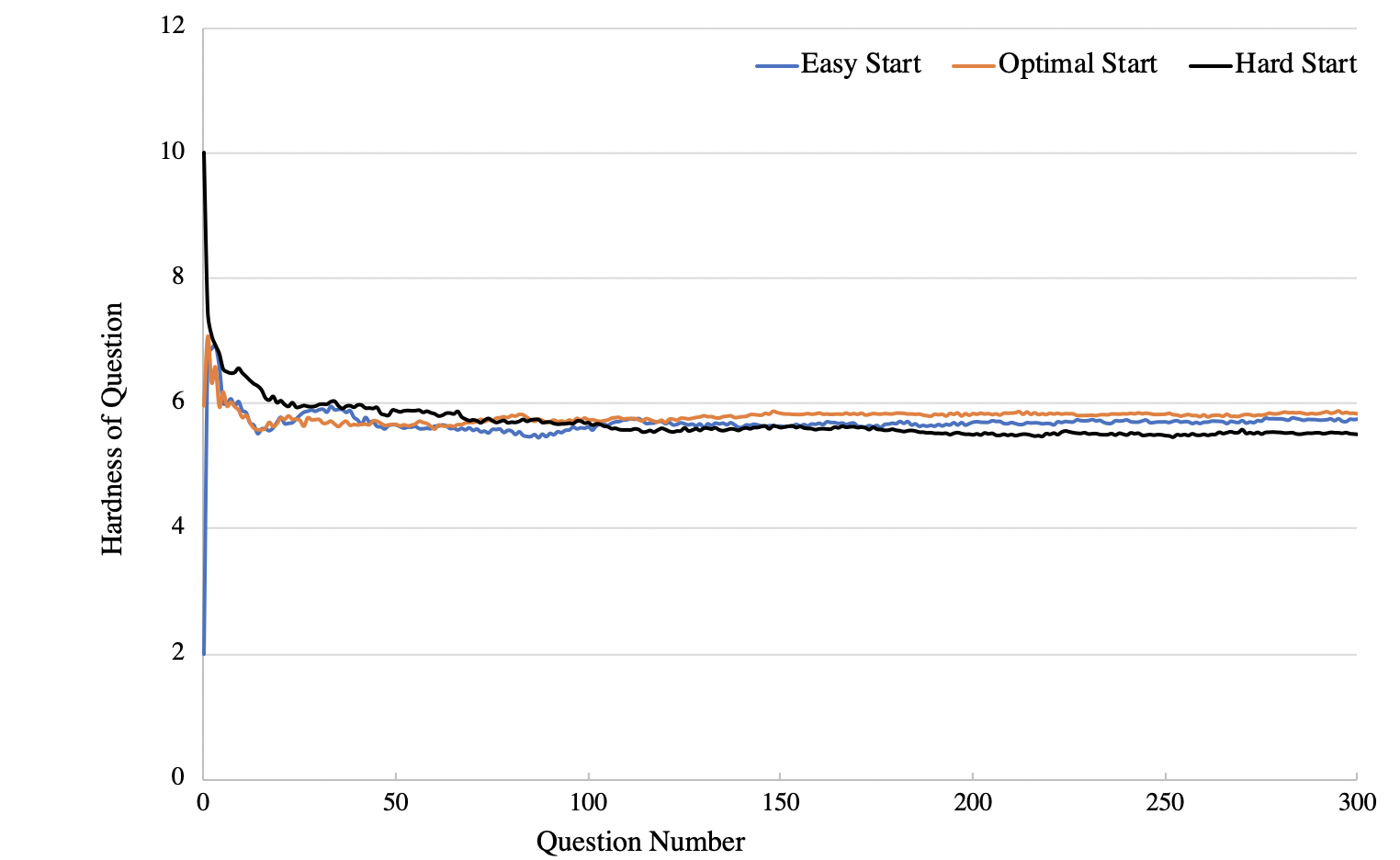}
\end{center}
\caption{Numerical illustration of inbulit exploration property for the algorithm. Hardness of question over time with three different starting point. }

\end{figure}


\begin{thebibliography}{1}


\bibitem{audibert2010best}
Jean-Yves Audibert and S{\'e}bastien Bubeck.
\newblock Best arm identification in multi-armed bandits.
\newblock In \emph{COLT-23th Conference on Learning Theory-2010}, pages 13--p,
  2010.

\bibitem{bartroff2008modern}
Jay Bartroff, Matthew Finkelman, and Tze~Leung Lai.
\newblock Modern sequential analysis and its applications to computerized
  adaptive testing.
\newblock \emph{Psychometrika}, 73\penalty0 (3):\penalty0 473--486, 2008.

\bibitem{bubeck2011pure}
S{\'e}bastien Bubeck, R{\'e}mi Munos, and Gilles Stoltz.
\newblock Pure exploration in finitely-armed and continuous-armed bandits.
\newblock 2011.

\bibitem{chernoff1959sequential}
Herman Chernoff.
\newblock Sequential design of experiments.
\newblock \emph{The Annals of Mathematical Statistics}, 30\penalty0
  (3):\penalty0 755--770, 1959.

\bibitem{even2006action}
Eyal Even-Dar, Shie Mannor, and Yishay Mansour.
\newblock Action elimination and stopping conditions for the multi-armed bandit
  and reinforcement learning problems.
\newblock \emph{Journal of machine learning research}, 7\penalty0
  (Jun):\penalty0 1079--1105, 2006.

\bibitem{garivier2016optimal}
Aur{\'e}lien Garivier and Emilie Kaufmann.
\newblock Optimal best arm identification with fixed confidence.
\newblock In \emph{Conference on Learning Theory}, pages 998--1027, 2016.

\bibitem{georgiadou2007review}
Elissavet~G Georgiadou, Evangelos Triantafillou, and Anastasios~A Economides.
\newblock A review of item exposure control strategies for computerized
  adaptive testing developed from 1983 to 2005.
\newblock \emph{The Journal of Technology, Learning and Assessment}, 5\penalty0
  (8), 2007.

\bibitem{juneja2019partition}
Sandeep Juneja and Subhashini Krishnasamy.
\newblock Sample complexity of partition identification using multi-armed
  bandits.
\newblock In Alina Beygelzimer and Daniel Hsu, editors, \emph{Proceedings of
  the Thirty-Second Conference on Learning Theory}, volume~99 of
  \emph{Proceedings of Machine Learning Research}, pages 1824--1852, Phoenix,
  USA, 25--28 Jun 2019. PMLR.
\newblock URL \url{http://proceedings.mlr.press/v99/juneja19a.html}.

\bibitem{kaufmann2016complexity}
Emilie Kaufmann, Olivier Capp{\'e}, and Aur{\'e}lien Garivier.
\newblock On the complexity of best-arm identification in multi-armed bandit
  models.
\newblock \emph{The Journal of Machine Learning Research}, 17\penalty0
  (1):\penalty0 1--42, 2016.

\bibitem{LaiRobbins85}
Lai, Tze Leung and Robbins, Herbert.
\newblock  Asymptotically efficient adaptive allocation rules.
\newblock \emph{Advances in applied mathematics}, 6\penalty0
  (1):\penalty0 4--22, 1985.


\bibitem{lewis1991computerized}
Charles Lewis, Kathleen~M Sheehan, Richard~N DeVore, and Leonard~C Swanson.
\newblock Computerized mastery testing system, a computer administered variable
  length sequential testing system for making pass/fail decisions, October~22
  1991.
\newblock US Patent 5,059,127.

\bibitem{lopez2007semi}
Marco L{\'o}pez and Georg Still.
\newblock Semi-infinite programming.
\newblock \emph{European Journal of Operational Research}, 180\penalty0
  (2):\penalty0 491--518, 2007.

\bibitem{magureanu2014lipschitz}
Stefan Magureanu, Richard Combes, and Alexandre Proutiere.
\newblock Lipschitz bandits: Regret lower bounds and optimal algorithms.
\newblock \emph{arXiv preprint arXiv:1405.4758}, 2014.

\bibitem{mannor2004sample}
Shie Mannor and John~N Tsitsiklis.
\newblock The sample complexity of exploration in the multi-armed bandit
  problem.
\newblock \emph{Journal of Machine Learning Research}, 5\penalty0
  (Jun):\penalty0 623--648, 2004.

\bibitem{naghshvar2013active}
Mohammad Naghshvar, Tara Javidi, et~al.
\newblock Active sequential hypothesis testing.
\newblock \emph{The Annals of Statistics}, 41\penalty0 (6):\penalty0
  2703--2738, 2013.

\bibitem{reckase1983procedure}
Mark~d Reckase.
\newblock A procedure for decision making using tailored testing.
\newblock In \emph{New horizons in testing}, pages 237--255. Elsevier, 1983.

\bibitem{van2010elements}
Wim~J van~der Linden and Cees~AW Glas.
\newblock \emph{Elements of adaptive testing}.
\newblock Springer, 2010.

\end{thebibliography}
\end{document}